%% file: main.tex
\documentclass[10pt,twocolumn,letterpaper]{article}

\usepackage{iccv}
\usepackage{times}
\usepackage{epsfig}
\usepackage{graphicx}
\usepackage{caption}
\usepackage{amsmath}
\usepackage{amssymb}
\usepackage{booktabs}
\usepackage{algorithm}
\usepackage{algorithmic}

\usepackage{indentfirst}
\usepackage{multirow}
\usepackage[accsupp]{axessibility}
\usepackage{enumitem}

\usepackage[pagebackref=true,breaklinks=true,letterpaper=true,colorlinks,bookmarks=false]{hyperref}

\iccvfinalcopy 

\def\iccvPaperID{9141} 
\def\httilde{\mbox{\tt\raisebox{-.5ex}{\symbol{126}}}}

\ificcvfinal\fi

\title{Partition-and-Debias: Agnostic Biases Mitigation via \\ A Mixture of Biases-Specific Experts}

\author{Jiaxuan Li \\
The University of Tokyo, Japan \\
{\tt\small li@nlab.ci.i.u-tokyo.ac.jp}
\and
Duc Minh Vo\\
The University of Tokyo, Japan\\
{\tt\small vmduc@nlab.ci.i.u-tokyo.ac.jp}
\and
Hideki Nakayama\\
The University of Tokyo, Japan\\
{\tt\small nakayama@ci.i.u-tokyo.ac.jp}
}

\begin{document}

\twocolumn[{%
\renewcommand\twocolumn[1][]{#1}%
\maketitle
\ificcvfinal\thispagestyle{empty}\fi
\begin{center}
    \centering
    \captionsetup{type=figure}
    \includegraphics[width=0.99\textwidth]{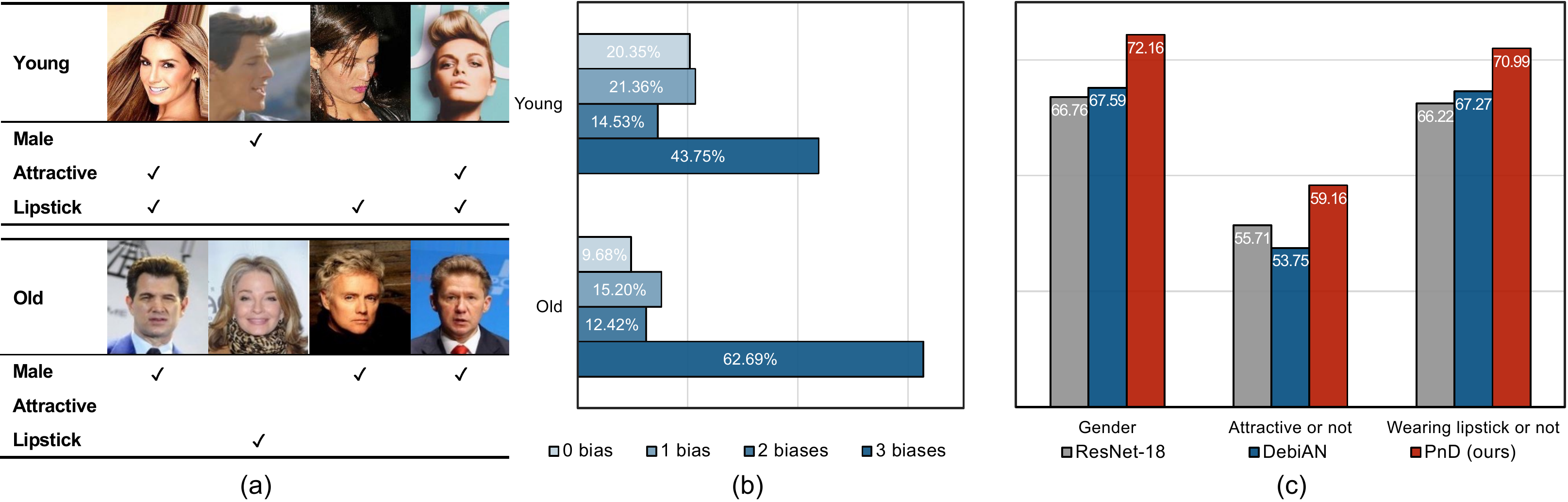}
    \captionof{figure}{Taking the age attribute in the CelebA dataset as an example for analyzing the agnostic biases problem. 
    (a) Representative samples in CelebA containing multiple biases. By analyzing the attribute distribution of all data within the \textit{young/old} category, we found that three attributes can be biased: gender (\textit{female/male}), attractiveness (\textit{attractive/not attractive}), and wearing lipstick (\textit{lipstick/no lipstick}). 
    (b) Proportion of samples with 0 -- 3 biases in a single image within the \textit{young} and \textit{old} groups.
    This indicates that the number of images with multiple biases dominated other cases in the dataset. 
    (c) Age classification accuracy (\%) of the existing methods for the worst groups of three bias attributes in CelebA degrades under this realistic bias scenario.
    For clarification, when discussing biases in this paper, we refer to abstract words like age as ``attribute”, and the \textit{italic} words which describe the labels of age like \textit{young}/\textit{old} as ``category”.
    } 
    \label{multiple_biases}
\end{center}%
}]


\begin{abstract}

Bias mitigation in image classification has been widely researched, and existing methods have yielded notable results. 
However, most of these methods implicitly assume that a given image contains only one type of known or unknown bias, failing to consider the complexities of real-world biases.
We introduce a more challenging scenario, \textbf{agnostic biases mitigation}, aiming at bias removal regardless of whether the type of bias or the number of types is unknown in the datasets.
To address this difficult task, we present the Partition-and-Debias (PnD) method that uses a mixture of biases-specific experts to implicitly divide the bias space into multiple subspaces and a gating module to find a consensus among experts to achieve debiased classification. 
Experiments on both public and constructed benchmarks demonstrated the efficacy of the PnD.
Code is available at: \url{https://github.com/Jiaxuan-Li/PnD}.

\end{abstract}


\section{Introduction}

One of the reasons for poor generalization in image classification is the presence of biased features in training data~\cite{Qi_2022_Class,Zhao_2017_Men,lee2021learning}, which distracts the model from learning the target features associated with the classification objects. 
Thus, accurately capturing the target features while reducing the influence of these biases\footnote{Similar to ~\cite{tartaglione2021end}, \textit{``bias"} refers to the attribute spuriously correlated with the target attribute.} has become a critical issue, resulting in increased bias mitigation research~\cite{Torralba_2011_Unbiased}.

Unlike most previous studies that implicitly assumed that only one type of known/unknown bias exists in a given image,
we investigate the coexistence of multiple unknown biases in an image.
For instance, most \textit{young} samples in CelebA~\cite{liu2015faceattributes} are associated with the \textit{female}, \textit{attractive}, and \textit{lipstick} categories (Fig.~\ref{multiple_biases}a), whereas the \textit{old} samples have corresponding yet reversed ones.
Consequently, for the age (\textit{young}/\textit{old}) classification in CelebA, these three biases, including gender, attractiveness, and wearing lipstick, degrade the prediction performance.
Overall, we discovered that 43.75\% of the \textit{young} samples had three biases, which means they were all annotated with \textit{female}, \textit{attractive}, and \textit{lipstick}, whereas 58.28\% of the samples had at least two of them (Fig.~\ref{multiple_biases}b).
The $\textit{old}$ samples show similar patterns.
These observations imply that multiple biases are inevitable in a given image. At the same time, we cannot determine all types of bias that may appear in the image.
Dealing with multiple unknown biases is thus emergent, and cannot be fully solved using prior methods (Fig.~\ref{multiple_biases}c) because (i) they fail to capture the biases of different types and (ii) removing a single bias does not always eliminate the effects of all biases.
Therefore, we introduce a more challenging scenario, \textbf{agnostic biases}, in which the unknown biases include not only the type of bias, but also the number of types.
Here, we use ``agnostic biases" to bring attention to biases in real-world scenarios, where the bias type and number of types are unknown.
We do not use ``unknown biases" proposed in~\cite{Jeon_2022_CVPR},
because it ignores multiple unknown biases.
Our scenario overcomes the existing bias constraints, boosting the performance of real-world applications.

We hypothesized and empirically found that the features of agnostic biases scatter at different depths of the network depending on the biases’ nature. 
Even if multiple biases are entangled at the same depth, they can be be regarded as one type of bias.
Thus, agnostic biases can be grouped by their feature levels and processed individually at different network depths. 
As a result, we propose a Partition-and-Debias (PnD) approach based on the divide-and-conquer strategy to capture and remove agnostic biases at different levels for debiased classification.
Thus, the entire agnostic bias scenario space is divided into multiple subscenario spaces that can be handled by multiple biases-specific experts.
The final prediction is obtained based on the consensus of all the experts using a gating module.
Our contributions are:

\begin{itemize}[leftmargin=*]

\item We point out the existence of multiple biases in the real world, proposing a new scenario with agnostic biases that fills in the gaps of previous works' bias assumptions.
 
\item We present a Partition-and-Debias approach to solve the new scenario via a mixture of biases-specific experts.

\item On both public and our constructed challenging bias datasets, experimental results show that the proposed method achieves cutting-edge performance.

\end{itemize}

\section{Related Work}

\subsection{Bias mitigation}

Bias mitigation learns the target features without influence by spurious correlations when training data is biased.

\noindent
\textbf{Known bias mitigation} assumes the annotation of bias or the type of bias is accessible.
The previous methods can be classified as supervised or unsupervised.
The former includes reweighting samples with higher uncertainty~\cite{li2019repair}, regularization~\cite{Sagawa*2020Distributionally,tartaglione2021end}, data augmentation~\cite{ramaswamy2021fair}, and supervised bias estimation~\cite{alvi2018turning,kim2019learning,adeli2021representation}.
The latter often uses mixup~\cite{hwang2022selecmix}, a two-branch network~\cite{nam2020learning,lee2021learning}, prioritizing simple target features while ignoring complex biased features~\cite{shrestha2022occamnets}, and MaskTune~\cite{asgari2022masktune}.
These methods make strong assumptions about the type of bias. 
For instance, bias can be easily learned ~\cite{nam2020learning,lee2021learning,hwang2022selecmix}, target features are simpler than bias features~\cite{shrestha2022occamnets}, and the bias is editable~\cite{asgari2022masktune}. 
Furthermore, most studies considered only a single type of bias appearing in an image. Although \cite{shrestha2022occamnets} used a multiple-bias dataset in their experiments, they still adhered to the limitations of the assumptions on the type of bias.
Our method belongs to the unsupervised approach, yet we relax the strong bias assumptions and use a partition-and-debias strategy.

\noindent
\textbf{Unknown bias mitigation} does not require a pre-definition for bias in the dataset. 
Jeon \etal~\cite{Jeon_2022_CVPR} proposed obtaining unbiased target features from the shallow layers of the classification network. However, their definition of unknown bias misses that there are multiple unknown biases in the data. For real-world datasets, spurious correlations are complex and cannot be defined simply as a result of a specific attribute. By contrast, our agnostic biases assumption emphasizes that both the type and number of bias types are unknown.
Also, Li \etal~\cite{li2022discover} proposed an Equal Opportunity Violation loss to discover the most salient bias from unknown biases and then mitigate it by reweighting. Although they considered two biases in their experiments, their method theoretically could only eliminate one dominating bias, which was binary rather than multi-class bias. In contrast to them, our model overcomes these limitations.

\subsection{Mixture of experts}
The mixture of experts (MoE) technique was originally proposed by Jacobs~\etal~\cite{jacobs1991adaptive} to mitigate the effects of different types of samples on the training data. It divides data into different domains using a gating network and assigns multiple experts to handle each domain. Recently, Zuo~\etal~\cite{zuo2022moebert} used MoE in language models by breaking a pretrained model into multiple experts to speed up the inference process. Zhang~\etal~\cite{Zhang_2019_ICCV} combined MoE with fine-grained categorization by training each subsequent expert using prior information obtained from the previous expert. 
Unlike these methods, we employ the MoE strategy for debiasing and specifically design it to remove agnostic biases by inserting experts at different depths of the network.

\section{Features of Different Levels Matters}
\label{bias_distribution}

\begin{figure}[tb]
\begin{center}
\centering\includegraphics[width=0.47\textwidth]{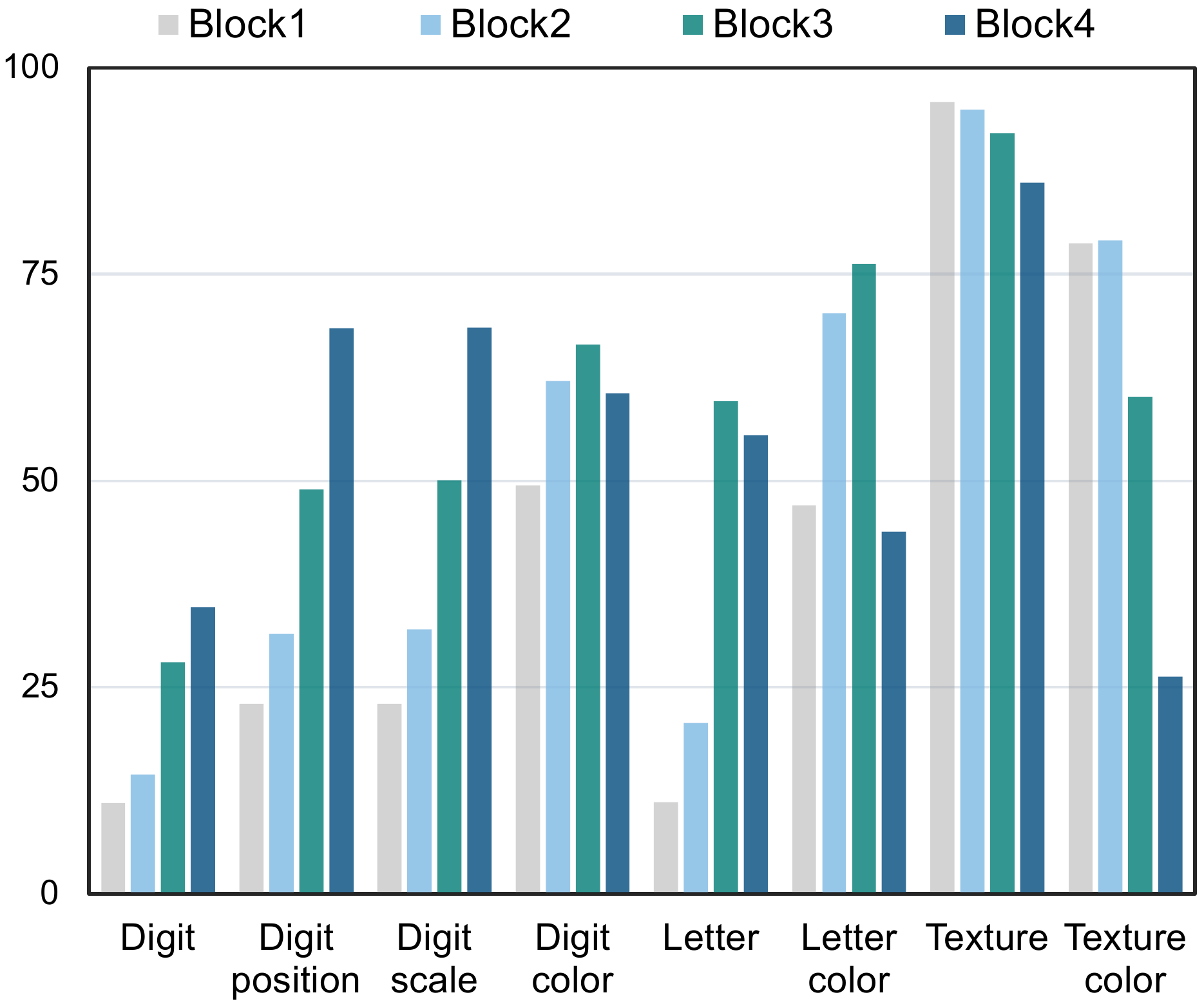}
\end{center}
\caption{Classification accuracy scores (\%) for the 8 attributes when retraining in features learned from target class classification at blocks of different depths in ResNet-18. We find that the classification performance for different attributes trends differently across the depths.} 
\label{bias_feature_dis}
\end{figure}

\noindent
\textbf{Our hypothesis.} 
When training a neural network with target categories, agnostic biases manifest as scattered features at different network depths.

\noindent
\textbf{Experimental setup.} 
We use the Biased MNIST~\cite{shrestha2022occamnets} dataset (see Sec.~\ref{sec_datasets}) in this exploratory experiment. 
The biases arise from the co-occurrence of each digit category with specific categories from all other attributes, such as digit color and digit position. 
Unlike an image with a single bias, one image in this dataset may have up to seven biases.
The bias ratio, which denotes the probability of co-occurrence, is 0.95. 
We selected ResNet-18~\cite{he2016deep} consisting of four residual blocks, as the classification network.

First, we trained a classification network from scratch using the target
categories \textit{0} -- \textit{9} in the digits and obtained an average classification accuracy of 33.73\% for all categories.
The learned features obtained from the trained model can then be visualized to investigate how bias features are distributed across the network when training the network with target categories.
However, since many attributes such as digital color, digit position, and digit scale are interdependent, their features overlap in feature maps, making it difficult to distinguish their differences by simply looking at feature maps. We used the classification accuracy for each attribute separately in each block to determine their distributions.
Specifically, we froze the trained network weights and added a binary classifier after each trained block. We trained the additional four classifiers to obtain the corresponding classification accuracies for all eight attributes.

\noindent
\textbf{Features of biases with different levels are distributed at different depths of the network.} 
We obtained $4 \times 8$ accuracy results after retraining, as shown in Fig.~\ref{bias_feature_dis}. 
(i) From the perspective of different attributes, the classification accuracy of all attributes except texture color was notably higher than that of the digit in the last block (block 4), which is usually used to determine the final prediction. Furthermore, the other blocks followed the same pattern as the last block. This phenomenon implies that the previously learned features from the target attribute classification (here, digits) are more easily classifiable in the bias attribute classification than in the target attribute classification.
We concluded that many spuriously correlated features exist at all depths of the network, degrading the target attribute predictions.
(ii) From the perspective of different blocks, although most bias attributes can be classified in each layer, the classification performance for some bias attributes varies depending on the block. Texture-relevant attribute classifiers performed well in the first block, while those with position- and scale-relevant attributes performed better in the last block; the remaining attributes achieved the best results in the third block. These findings are consistent with our intuition regarding the distribution of image features, which holds that texture features are more abundant in the shallower parts of the network and that spatial and scale information are more prevalent in the deeper parts of the network.
We conclude that each bias attribute feature exists at all network depths, yet these features are clustered at different network depths.

\begin{figure*}[ht!]
\centering\includegraphics[width=0.98
\textwidth]{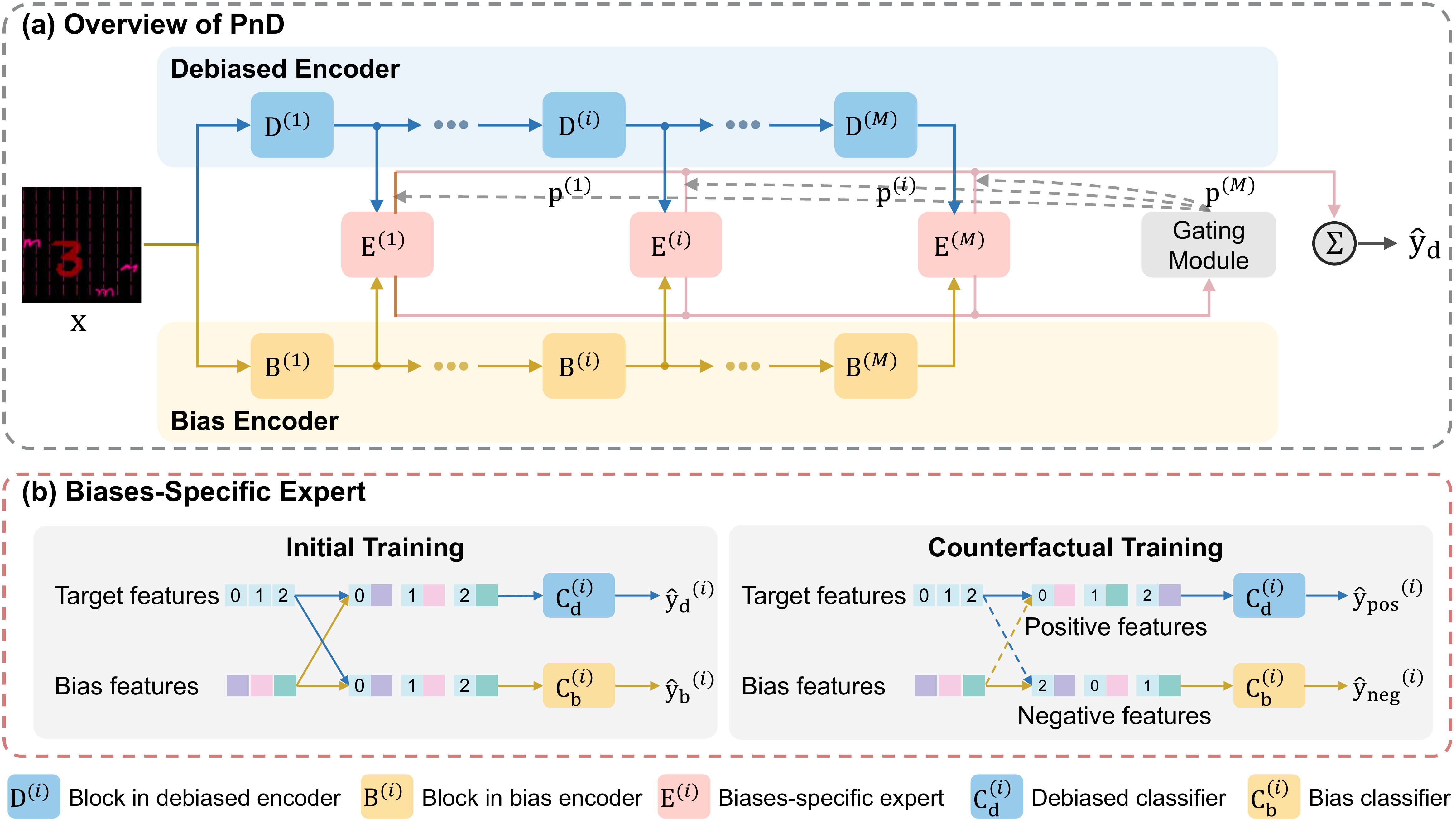}
\caption{(a) Schematic of our proposed PnD, including a debiased encoder, a bias encoder, multiple biases-specific experts, and a gating module. The debiased encoder and bias encoder extract target and bias features from input images, which are fed into the biases-specific expert after each block for debiased classification and bias detection. The gating module adaptively mixes all the debiased classification results for the final output. (b) Biases-specific expert. It processes target features and bias features by combining them in the order of the original input batch for initial training, and recombining them to generate positive and negative samples during counterfactual training.}
\label{fig2}
\end{figure*}

\section{Proposed Partition-and-Debias}

The above experiment suggests that an ideal strategy for resolving our problem should be able to remove as many biases from the network depths as possible.
Thus, we adopt a partition-and-debias strategy in our method, namely PnD, which divides the entire agnostic bias scenario space into different subscenario spaces across the classification network depths.
Multiple biases of the same level are allowed in each subscenario space because they can be viewed naturally as a type of bias.
This simple concept overcomes the limitations of previous studies allow the model to simultaneously capture and remove multiple biases at one time.

Our PnD consists of a debiased encoder $\mathcal{D}$, bias encoder $\mathcal{B}$, biases-specific experts $\mathcal{E}$, and gating module (Fig.~\ref{fig2}).
Both $\mathcal{D}$ and $\mathcal{B}$ contain several blocks of convolution layers to generate the target and bias features from an image (Sec.~\ref{sub_encode}).
$\mathcal{E}$ are responsible for purifying the target and bias features under agnostic bias scenarios (Sec.~\ref{sub_expert}).
Finally, a gating module adaptively gathers all the expert predictions before making a final decision (Sec.~\ref{sub_gate}).

\subsection{Target and bias features extraction} 
\label{sub_encode}
Given an image $\mathbf{x}$ with the target label $\mathbf{y}$ (vector-like is used to represent class $y$ for simplicity), we use a debiased encoder $\mathcal{D}=\{ D^{(i)} \}_{i=1}^M$ and bias encoder $\mathcal{B}=\{B^{(i)} \}_{i=1}^M$ to extract target and bias features separately.
Note that $D^{(i)}$ and $B^{(i)}$ are residual blocks in ResNet~\cite{he2016deep} although the network architecture can be any and $M=4$.

The image is fed into $\mathcal{D}$ to obtain the target features $\mathbf{z}_{\rm d}^{(i)}$ in $i^{\rm th}$ block.
Simultaneously, we obtain the bias features $\mathbf{z}_{\rm b}^{(i)}$ for each block in $\mathcal{B}$.
The size of $\mathbf{z}_{\rm d}^{(i)}$ is identical to that of $\mathbf{z}_{\rm b}^{(i)}$.
We omitted the feature size when referencing the extracted features to simplify the process.
Next, biases-specific experts processed these features.

\subsection{Biases-specific experts} 
\label{sub_expert}

The biases-specific experts $\mathcal{E}$ consist of four experts $E^{(i)}$.
Each of them contains two classifiers: a debiased classifier $C_{\rm d}^{(i)}$ and bias classifier $C_{\rm b}^{(i)}$.
The inputs of each $E^{(i)}$ are created from the features $\mathbf{z}_{\rm d}^{(i)}$ and $\mathbf{z}_{\rm b}^{(i)}$ obtained from the corresponding $D^{(i)}$ and $B^{(i)}$.
We combined $\mathbf{z}_{\rm d}^{(i)}$ and $\mathbf{z}_{\rm b}^{(i)}$ features in two ways, creating the original and counterfactual features used in our two-stage training (initial and counterfactual trainings). In both training stages, debiased classifier $C_{\rm d}^{(i)}$ and bias classifier $C_{\rm b}^{(i)}$ are used for {debiased classification and bias detection, respectively.

\subsubsection{Initial training}
We combine the features $\mathbf{z}_{\rm d}^{(i)}$ and $\mathbf{z}_{\rm b}^{(i)}$
to create the original features $\mathbf{z}^{(i)}=[\mathbf{z}_{\rm d}^{(i)}; \mathbf{z}_{\rm b}^{(i)}]$ ($[\cdot;\cdot]$ denotes concatenation) (Fig.~\ref{fig2}b, left). 
The $i^{th}$ expert $E^{(i)}$ takes $\mathbf{z}^{(i)}$ as the input, and outputs a bias detection result $\mathbf{\hat{y}}_{\rm b}^{(i)}$ and a debiased classification result $\mathbf{\hat{y}}_{\rm d}^{(i)}$  made by $C_{\rm b}^{(i)}$ and $C_{\rm d}^{(i)}$, respectively.

\noindent
\textbf{Bias detection.} 
This encourages the bias encoder to learn the bias features.
Because the bias features are easier to learn during training with target categories, we can concentrate our bias encoder on the more easily learned features by employing $\textsc{GCE}$ loss \cite{r6}
as discussed in~\cite{nam2020learning,lee2021learning,Kim_2021_BiaSwap}, although the bias information in the dataset is unavailable:

\begin{equation}
\mathcal{L}_{\textup{bias}} =\sum_{i=1}^{M}\textsc{GCE}\left(\mathbf{\hat{y}}_{\rm b}^{(i)}, \mathbf{y}\right).
\label{L_bias}
\end{equation}

\noindent
\textbf{Debiased classification.}
To optimize the debiased and bias encoders separately, as opposed to bias detection, debiased classification should prioritize unbiased samples, which do not contain bias features and are difficult to fit using the bias encoder.
Consequently, when bias detection is used, these samples are misclassified or classified with lower confidence, whereas debiased classification classifies them correctly or with higher confidence.
Considering this, we follow \cite{nam2020learning} and add a weight ${\rm w}^{(i)}$ to each sample in debiased classification. ${\rm w}^{(i)}$ is defined as: ${\rm w}^{(i)} = \frac{\textsc{CE}\left(\mathbf{\hat{y}}_{\rm b}^{(i)}, \mathbf{y}\right)}{\textsc{CE}\left(\mathbf{\hat{y}}_{\rm d}^{(i)}, \mathbf{y}\right)+\textsc{CE}\left(\mathbf{\hat{y}}_{\rm b}^{(i)}, \mathbf{y}\right)}$,
where we use $\textsc{CE}(\mathbf{\hat{y}}_{\rm d}^{(i)}, \mathbf{y})$ and $\textsc{CE}(\mathbf{\hat{y}}_{\rm b}^{(i)}, \mathbf{y})$ to measure the relative difficulty between debiased classification and bias detection, $\textsc{CE}(\cdot, \cdot)$ denotes cross-entropy loss function. The loss for debiased classification is expressed as:

\begin{equation}
\mathcal{L}_{\textup{debias}} =\sum_{i=1}^{M}  {\rm w}^{(i)} \times \textsc{CE}\left(\mathbf{\hat{y}}_{\rm d}^{(i)}, \mathbf{y}\right).
\label{L_debias}
\end{equation}

Combining Eq.~\ref{L_bias} and Eq.~\ref{L_debias}, we obtain the total classification loss $\mathcal{L}_{\textup{cls}}$ for debiased classification and bias detection: $\mathcal{L}_{\textup{cls}} = \alpha \times \mathcal{L}_{\textup{debias}} + \mathcal{L}_{\textup{bias}}$,
$\alpha$ is hyperparameter that balances $\mathcal{L}_{\textup{debias}}$ and $\mathcal{L}_{\textup{bias}}$;
$\mathcal{L}_{\textup{debias}}$ forces the debiased classification to focus more on unbiased samples with weight ${\rm w}^{(i)}$ added to \textsc{CE} loss, whereas
$\mathcal{L}_{\textup{bias}}$ focuses on bias features owing to the \textsc{GCE} loss to support easier-learned features.

\noindent
\textbf{Diversity penalty for biases-specific experts.} 
To achieve diversified biases-specific experts, we introduc a Kullback-Leibler (KL) divergence-based loss function~\cite{wu2021r} to penalize the bias detection of each expert. The diversity loss for experts can be formulated as:

\begin{equation}
\mathcal{L}_{\textup{div}}=\sum_{i=2}^{M} \textup{exp}\left(-\textsc{KL}\left(\mathbf{\hat{y}}_{\rm b}^{(i)}, \mathbf{\hat{y}}_{\rm b}^{(i-1)}\right)\right).
\label{L_div}
\end{equation}

\noindent 
Thus, using Eq.~\ref{L_div}, we can regularize the diversity of the bias detection by each expert, allowing them to capture as many biases as possible. In this way, each expert can focus on different level features, and thus, different biases.

\subsubsection{Counterfactual training} 
We obtain relatively accurate bias and target features after warming the model during the initial training. Counterfactual training is used to further separate target features from bias features.
This approach is based on two counterfactual procedures.
(i) When we change the sample's target features while keeping its bias features unchanged, the model's decision should be changed; (ii) When we keep its target features unchanged while changing the sample's bias features, the model should make the same decision for the changed features as for the original features. To leverage these two procedures, we first synthesize counterfactual features before conducting counterfactual inference using contrastive loss.

\noindent
\textbf{Synthesizing counterfactual features.} 
We randomly sample a mini-batch of $K$ samples to construct the counterfactual features. 
For the $j^{th}$ sample,
in the mini-batch, we first randomly select one bias feature and $P$ target features from the other samples as follows: 
${\mathcal{\tilde{Z}}_{\rm b}}^{(i)}=\{{\mathbf{z}}_{{\rm b}_{q}}^{(i)}\}$, and ${\mathcal{\tilde{Z}}_{\rm d}}^{(i)}=\{{\mathbf{z}}_{{\rm d}_{l}}^{(i)}\}_{l=1}^{P}$, where $q \neq j$, $l \neq j$, and $0<q\leq{K}$.

Subsequently, the target feature $\mathbf{z}_{{\rm d}_{j}}^{(i)}$ is paired with the selected bias feature to construct positive features ${{\mathcal{Z}}_{\textup{pos}}}^{(i)}=\{[{\mathbf{z}_{{\rm d}_{j}}^{(i)}}; {\mathbf{z}}_{{\rm b}_{q}}^{(i)}]\}$ (Fig.~\ref{fig2}b, right). Similarly, the bias feature ${\mathbf{z}_{{\rm b}_{j}}^{(i)}}$ is paired with the other $P$ target features to construct its negative features  ${{\mathcal{Z}}_{\textup{neg}}}^{(i)}=\{[{\mathbf{z}}_{{\rm d}_{l}}^{(i)}; {\mathbf{z}_{{\rm b}_{j}}^{(i)}}]\}_{l=1}^{P}$.

\noindent
\textbf{Counterfactual inference.}
Positive and negative features were fed into the debiased classifier $C_{\rm d}^{(i)}$ and bias classifier $C_{\rm b}^{(i)}$, respectively. We obtain a positive prediction $\mathcal{Y}_{\textup{pos}}^{(i)}=\{{\mathbf{\hat{y}}_{\textup{pos}}}^{(i)}\}$ and a set of negative predictions $\mathcal{Y}_{\textup{neg}}^{(i)}=\{{\mathbf{\hat{y}}_{\textup{neg}_{l}}}^{(i)}\}_{l=1}^{P}$ for the original result $\mathbf{\hat{y}}_{\rm d}^{(i)}$.
We then use the contrastive loss $\mathcal{L}_{\textup{con}}$ for this counterfactual inference:

\begin{equation}
\nonumber
\mathcal{L}_{\textup{con}}= \sum_{i=1}^{M} -\textup{log}\frac{\textup{exp} \left(-\textup{dist}\left(\mathbf{\hat{y}}_{\rm d}^{(i)}, \mathbf{\hat{y}}_{\textup{pos}}^{(i)}\right)\right)}{\sum_{\mathbf{y}^{\prime} \in \mathcal{Y}_{\textup{neg}}^{(i)} \cup \{{\mathbf{\hat{y}}_{\textup{pos}}}^{(i)}\}} \textup{exp} \left(-\textup{dist}\left(\mathbf{\hat{y}}_{\rm d}^{(i)}, \mathbf{y}^{\prime}\right)\right)},
\end{equation}

\noindent
where $\textup{dist}(\cdot,\cdot)$ denotes Euclidean distance. 
This encourages the model to group samples with identical target features into the same category, regardless of their bias features. Conversely, even if samples have the same bias features, they can be classified into different categories if they have different target features.

\subsection{Mixture of biases-specific experts using adaptively gating} 
\label{sub_gate}
The final output $\mathbf{\hat{y}}_{\rm d}$ of the model is obtained by combining the debiased classification results $\mathbf{\hat{y}}_{\rm d}^{(i)}$ from each biases-specific expert through an gating module. The gating loss for this operation can be presented as: $\mathcal{L}_{\textup{gate}}=\textsc{CE}\left(\mathbf{\hat{y}}_{\rm d}, \mathbf{y}\right)$.
$\mathbf{\hat{y}}_{\rm d} =\sum_{i=1}^M \rm {p^{(i)}} \times \mathbf{\hat{y}}_{\rm d}^{(i)}$, 
where $\rm {p^{(i)}}$ denotes the probability value assigned to the biased classification result $\mathbf{\hat{y}}_{\rm d}^{(i)}$ of $E^{(i)}$; it is the softmax result from the gating module by taking all experts’
debiased classification results as the input.
We call this module ``gating” derived from the ``gating" in MoE~\cite{jacobs1991adaptive}, where it refers to the weighted inputs from the gating network followed by a softmax function.

The complete loss for updating the entire model is:

\begin{equation}
\label{loss_all}
\mathcal{L} = \left\{\begin{array}{ll}
\mathcal{L}_{\textup{cls}} + \mathcal{L}_{\textup{gate}} + \mathcal{L}_{\textup{div}}& \text {initial training} \\
\mathcal{L}_{\textup{cls}} + \mathcal{L}_{\textup{gate}} + \mathcal{L}_{\textup{div}}  + \beta \times \mathcal{L}_{\textup{con}} & \text {counterfactual training}
\end{array}\right.
\end{equation}

\noindent 
where $\beta$ balances $\mathcal{L}_{\textup{con}}$ with other terms.

\section{Experiments}

\subsection{Datasets}
\label{sec_datasets}

\noindent
\textbf{Biased MNIST}~\cite{shrestha2022occamnets} contains ten digits (\textit{0} -- \textit{9}) as its target categories and seven biases: digit color, digit scale, digit position, type of background texture, background texture color, co-occurring letter, and letter color.
There are 50000, 10000, 10000 images for training, validation, and testing.

\noindent
\textbf{BAR}~\cite{nam2020learning} consists of typical action-place pairs, like \textit{climbing} and \textit{rockwall} in the training set; and unseen samples beyond the settled pairs in the test set.
There are six target actions in 1941 training and 654 test images.

\noindent
\textbf{Modified IMDB} is our constructed dataset using IMDB face images~\cite{Rasmus_2018_Deep}, containing 20000 training, 1617 validation, and 1617 test images.
The targets are \textit{young} and \textit{old}, and the biases are gender and wearing glasses (Fig.~\ref{fig_bface}).

\noindent
\textbf{MIMIC-CXR + NIH}
was constructed by simulating the biases brought about by different data sources when collecting the datasets. We mixed the MIMIC-CXR~\cite{Johnson_2019_MIMIC} and NIH~\cite{Wang_2017_Chestx} datasets into a MIMIC-CXR + NIH dataset. 
The target categories are \textit{no finding} and \textit{pneumonia}, and the biases come from two data sources where the correlation between the target and biases is not tangible. It contains 8500 training, 500 validation, and 500 test images.

\begin{figure}[t]
\begin{center}
\centering\includegraphics[width=0.45\textwidth]{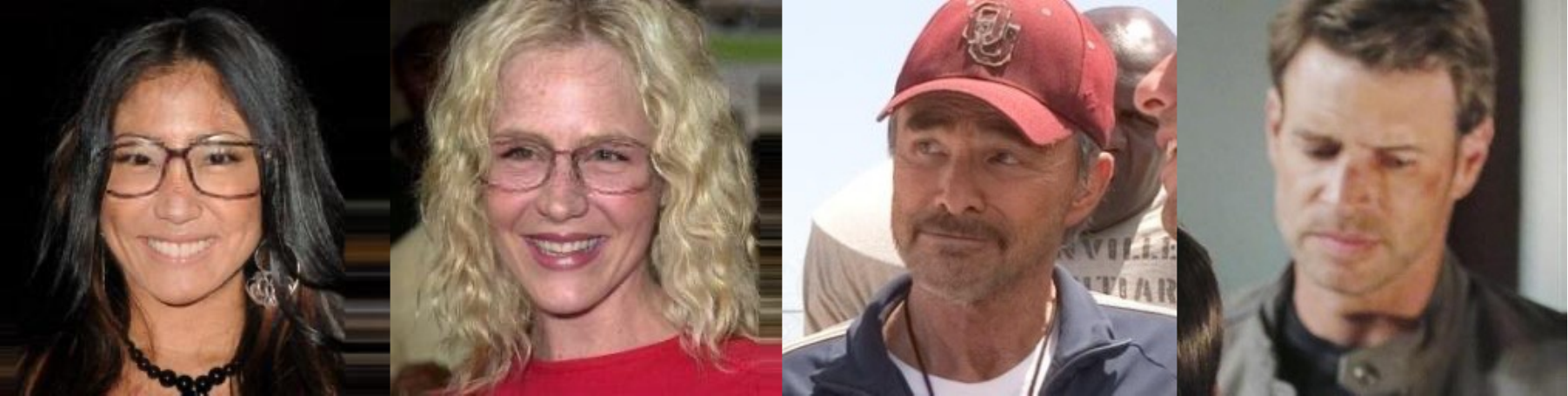}
\end{center}
\caption{Examples from Modified IMDB dataset, the left two are annotated with \textit{young}, but also with \textit{female} and \textit{wearing glasses}. The right two are annotated with \textit{old}, but also with \textit{male} and \textit{not wearing glasses}.
}
\label{fig_bface}
\end{figure}

\subsection{Implementation details}

\noindent
\textbf{Model architecture.} We employed feature extraction layers of ResNet-18~\cite{he2016deep} as the backbone of the debiased encoder and bias encoder. Two convolutional and two linear layers were used to design classifiers for biases-specific experts, and one linear layer was used to construct the gating module. 

\noindent
\textbf{Training procedure.} Our PnD was built using PyTorch, and all the experiments were conducted on an NVIDIA RTX A4000 GPU. 
For input to the PnD model, all images were resized to 160 × 160 × 3 except for the BAR where they were randomly cropped to $224 \times 224 \times 3$ and horizontally flipped following~\cite{nam2020learning}.

\begin{table*}[tb]
{\centering
\caption{Accuracy scores (\%) on Biased MNIST, Modified IMDB, MIMIC-CXR + NIH, and BAR datasets with different bias ratios. We compare our proposed method with ResNet-18 and other SOTA methods. Our method is clearly far superior or close to other methods. The best results are highlighted in \textcolor{blue}{\textbf{blue}}, and the second-best results are in \textcolor{red}{\textbf{red}}.} 
\label{tab_overall_comparison}
\resizebox{\linewidth}{!}{
\begin{tabular}{l|cc|c|cc|cc}
\toprule
\multirow{2}{*}{Method} & \multicolumn{2}{c|}{Biased MNIST (7 biases)} & BAR (1 bias) & \multicolumn{2}{c|}{Modified IMDB (2 biases)} & \multicolumn{2}{c}{MIMIC-CXR + NIH (1 bias)}                 \\ 
& 0.75 & 0.95 & & 0.95 & 0.99 & 0.80 & 0.95 \\
\midrule
ResNet-18~\cite{he2016deep} &94.49 $\pm$ 0.15  &53.86 $\pm$ 1.90  & 51.85 $\pm$ 5.92 &\textcolor{red}{\textbf{74.31}} $\pm$ 0.44 &\textcolor{blue}{\textbf{67.04}} $\pm$ 2.07 & 64.53 $\pm$ 2.07 &56.83 $\pm$ 2.07 \\
LfF~\cite{nam2020learning} & 84.58 $\pm$ 2.46 & 35.16 $\pm$ 9.80 & 62.98 $\pm$ 2.76 &63.22 $\pm$ 1.53 &62.01 $\pm$ 1.58 & 55.40 $\pm$ 0.00 & 56.23 $\pm$ 0.67 \\
DFA~\cite{lee2021learning}   & 90.79 $\pm$ 0.14 & 44.52 $\pm$ 2.51 & 58.97 $\pm$ 1.28 & 64.19 $\pm$ 2.92  & 62.46 $\pm$ 0.71 & 52.67 $\pm$ 2.70  & 50.93 $\pm$ 0.58          \\
OccamNet~\cite{shrestha2022occamnets} & \textcolor{red}{\textbf{96.06}} $\pm$ 0.33 & \textcolor{red}{\textbf{66.85}} $\pm$ 0.55 & 52.60 $\pm$ 1.90 & 68.17 $\pm$ 0.99 & 61.60 $\pm$ 1.07 & 61.93 $\pm$ 0.40 & 52.15 $\pm$ 0.35  \\
DebiAN~\cite{li2022discover} & 90.90 $\pm$ 1.36 & 46.52 $\pm$ 2.65   & \textcolor{blue}{\textbf{69.88}} $\pm$ 2.92 & 72.42 $\pm$ 0.33 & 65.99 $\pm$ 0.80   & \textcolor{red}{\textbf{67.40}} $\pm$ 0.96 & \textcolor{red}{\textbf{60.00}} $\pm$ 1.40 \\
UBNet~\cite{Jeon_2022_CVPR}  &90.40 $\pm$ 0.05    & 54.31 $\pm$ 1.13 & 61.93 $\pm$ 0.46 &70.62 $\pm$ 0.25 &63.02 $\pm$ 0.21   &66.00 $\pm$ 0.46 &55.00 $\pm$ 0.17  \\ 
\midrule
PnD  & \textcolor{blue}{\textbf{96.60}} $\pm$ 0.22   & \textcolor{blue}{\textbf{70.43}} $\pm$ 0.74     &\textcolor{red}{\textbf{69.83}} $\pm$ 2.09    &\textcolor{blue}{\textbf{74.34}} $\pm$ 0.22   &\textcolor{red}{\textbf{66.58}} $\pm$ 0.26        & \textcolor{blue}{\textbf{67.87}} $\pm$ 0.91  & \textcolor{blue}{\textbf{60.73}} $\pm$ 0.87 \\     
\bottomrule
\end{tabular}
}
}
\end{table*}

\section{Results and Analysis}

We compared our model to ResNet-18~\cite{he2016deep}, LfF~\cite{nam2020learning}, DFA~\cite{lee2021learning}, OccamNet~\cite{shrestha2022occamnets}, DebiAN~\cite{li2022discover}, and UBNet~\cite{Jeon_2022_CVPR}. 
ResNet-18 was pretrained using ImageNet~\cite{deng2009imagenet} and simply used cross-entropy as its loss function without any debiasing strategy.
In all the experiments, we calculated the means and standard deviations of accuracies of the test set across three runs for all datasets. Unless otherwise specified, the bias ratio is 0.95 for all cases in the following subsections.

\subsection{Comparisons against state-of-the-art}

\noindent
\textbf{Overall comparisons.}
We report the accuracy scores of all compared methods in Tab.~\ref{tab_overall_comparison}.
We used the results with two different bias ratios for each dataset except for BAR, because 
almost all of its training images were biased, and no bias labels were provided.
We selected a relatively large bias ratio and a small bias ratio for this set of experiments. 
Nevertheless, due to the limitations of the original dataset, we could not set a smaller bias ratio for the Modified IMDB.

We can see that PnD outperforms all methods on Biased MNIST and MIMIC-CXR + NIH.
Meanwhile, for the BAR, our method achieved the second-best performance, which is comparable to the results of DebiAN~\cite{li2022discover}.
BAR has only one type of bias for each target category, and the images in the training set are purely biased. In addition, the proposed framework requires unbiased data. Therefore, our accuracy score is slightly lower than that of DebiAN~\cite{li2022discover}. For the Modified IMDB, PnD contains the best or second-best scores. Because there are only two biases in this dataset and this classification task is relatively simpler than that of the Biased MNIST, ResNet-18 also works well on it, whereas the SOTAs are inferior to PnD and ResNet-18 on this dataset. We conclude that PnD performs best in agnostic biases mitigation owing to the mixture of biases-specific experts, especially in the presence of multiple biases. Even when the number of biases was small, its performance was comparable to that of the others.

\begin{figure}[t]
\begin{center}
\centering\includegraphics[width=0.47\textwidth]{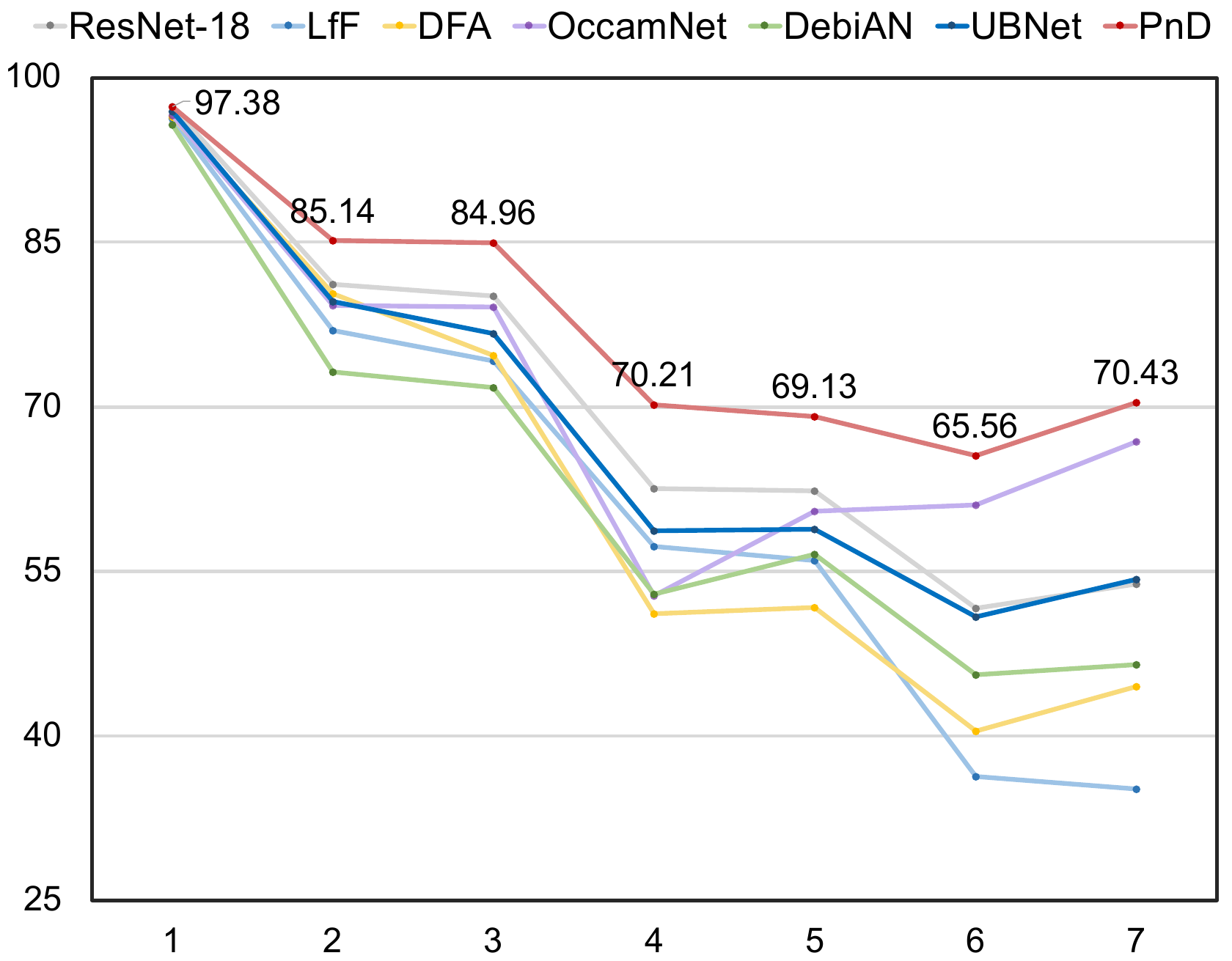}
\end{center}
\caption{The accuracy scores (\%) when changing the number of bias types from 1 to 7 in Biased MNIST. Regardless of the number of biases, PnD always achieves the best.} 
\label{num_bias}
\end{figure}

\noindent
\textbf{Robustness to different numbers of bias types.}
To evaluate the performance of all methods under different numbers of biases, we synthesized multiple biased MNISTs with varying numbers of biases (ranging from 1 to 7) by gradually adding digit color, digit scale, digit position, texture, texture color, letter, and letter color following the data synthesis operation in~\cite{shrestha2022occamnets} as shown in the supplement.} 

Regardless of the number of biases, PnD always achieved the best performance (Fig.~\ref{num_bias}).
When the number of biases was one (digit color), all methods achieved high scores.
This is because, at this time, the digit only occupies a small area in the center of the images, resulting in difficulty in learning digit color features.
Owing to the partition-and-debiasing strategy, our method does not eliminate 
its performance was as fast as those of the other methods after adding the second bias. Although our method suffers from a performance drop after the fourth bias, it still outperforms the other methods and remains nearly stable when additional biases are added.

\subsection{Ablation study}

\begin{table}[tb]
{\centering
\caption{Ablation study on Biased MNIST and MIMIC-CXR + NIH, including ablating the loss terms in PnD (3rd -- 6th rows), the implementation strategies (7th -- 8th rows), and training procedure (9th -- 10th rows). The results reveal that each component in PnD is effective.}
\label{ablation}
\resizebox{\linewidth}{!}{
\begin{tabular}{cccccc}
\toprule
\multicolumn{4}{c}{Method} & \multicolumn{2}{c}{Dataset} \\ 
\cmidrule(lr){1-4} \cmidrule(lr){5-6}{$\mathcal{L}_{\textup{cls}}$} & \multicolumn{1}{c}{$\mathcal{L}_{\textup{gate}}$}  & \multicolumn{1}{c}{$\mathcal{L}_{\textup{div}}$} & \multicolumn{1}{c}{$\mathcal{L}_{\textup{con}}$}  & Biased MNIST & MIMIC-CXR + NIH \\ 
\midrule
\checkmark &   &   &   &44.11 $\pm$ 0.76   &54.30 $\pm$ 0.46  \\ 
\checkmark &\checkmark   &  &   &68.90 $\pm$ 0.38   &57.87 $\pm$ 0.71   \\ 
\checkmark  &\checkmark   &\checkmark   &   & 69.88 $\pm$ 1.43  &58.10 $\pm$ 0.99    \\ 
\checkmark &\checkmark   &   &\checkmark   &69.48 $\pm$ 1.78   & 59.20 $\pm$ 0.26  \vspace{-0.6em}  \\ 
\multicolumn{6}{c}{\dotfill} \\

\multicolumn{4}{l}{w/o concatenation}    &67.11 $\pm$ 0.90  &58.57 $\pm$ 0.49  \\
\multicolumn{4}{l}{w/o adaptive gating} &67.84 $\pm$ 0.38  &58.13 $\pm$ 0.64  \vspace{-0.6em} \\
\multicolumn{6}{c}{\dotfill} \\

\multicolumn{4}{l}{initial training only}     &69.03 $\pm$ 0.53  &58.07 $\pm$ 1.27    \\ 
\multicolumn{4}{l}{counterfactual training only}      &69.61 $\pm$ 1.22   &59.00 $\pm$ 0.56  \vspace{-0.6em} \\ 
\multicolumn{6}{c}{\dotfill} \\
\multicolumn{4}{l}{PnD (full model) }     & 70.43 $\pm$ 0.74     & 60.73  $\pm$ 0.87  \\ 

\bottomrule
\end{tabular}
}
}
\end{table}

\noindent
\textbf{Ablation study for different loss terms.}
The ablation study results on the biased MNIST and MIMIC-CXR + NIH datasets are shown in Tab.~\ref{ablation}. We evaluated the impact of using multiple loss terms in Eq.~\ref{loss_all} by dropping each loss term individually (3rd -- 6th rows).
Note that we drop each loss term in both training phases. For clarity, we only discuss the impact of the loss on the overall framework, not involving the analysis of the two training phases.
The model with only $\mathcal{L}_{\textup{cls}}$ (3rd row) (i.e., the model with only two encoders and two classifiers following the ends of the blocks) performed the worst.
This is because it attempted to remove agnostic biases only once from the end of the network, as in previous studies, ignoring the fact that the number and type of agnostic biases are unknown.
When other loss terms are gradually added, the performance improves.
Particularly, the model with $\mathcal{L}_{\textup{gate}}$ (4th row) boosts the performance significantly because we begin to process the agnostic biases according to the network depth.
Moreover, the model with either $\mathcal{L}_{\textup{div}}$ or $\mathcal{L}_{\textup{con}}$ (5th and 6th rows) slightly improves the performance (less than 1\%).
When we used both $\mathcal{L}_{\textup{div}}$ and $\mathcal{L}_{\textup{con}}$ (11th row), the performance increased by 1.53\% compared to the model with $\mathcal{L}_{\textup{gate}}$. 
This is because $\mathcal{L}_{\textup{div}}$ regularizes the diversity of each block in the bias encoder, which also increases the diversity of counterfactual features, thus improving the results.
We conclude that the absence of any loss term reduces PnD's performance, indicating that all terms work properly and contribute to the final results.

\noindent
\textbf{Ablation study for different strategies used in our implementation.}
In PnD, we do not feed the bias and target features into the classifier separately for classification, but combine them and then classify them. The purpose, on the one hand, is to make the bias features a perturbation term on the target features to prevent overfitting, and on the other hand, to facilitate the synthesis of counterfactual features. We verified the results of a separate classification in our ablation experiments without concatenation (7th row), which showed significant decreases of 3.32\% and 2.16\%. For the adaptively gating operation, we demonstrated the effect of unweighted averaging (8th row), with a drop of 2.59\% and 2.60\% without adaptive gating.

\noindent
\textbf{Ablation study for two training phases.}
For a fair comparison, we still keep the same epochs. 
We can see that the two-stage training is better than only the first stage (9th row) or the second stage (10th row).
The first training provides relatively purer target and bias features for the second stage, whereas the second stage further disentangles these two features via counterfactual inference. Consequently, performance improves when the two stages work together.

\begin{figure}[tb]
\begin{center}
\centering\includegraphics[width=0.42\textwidth]{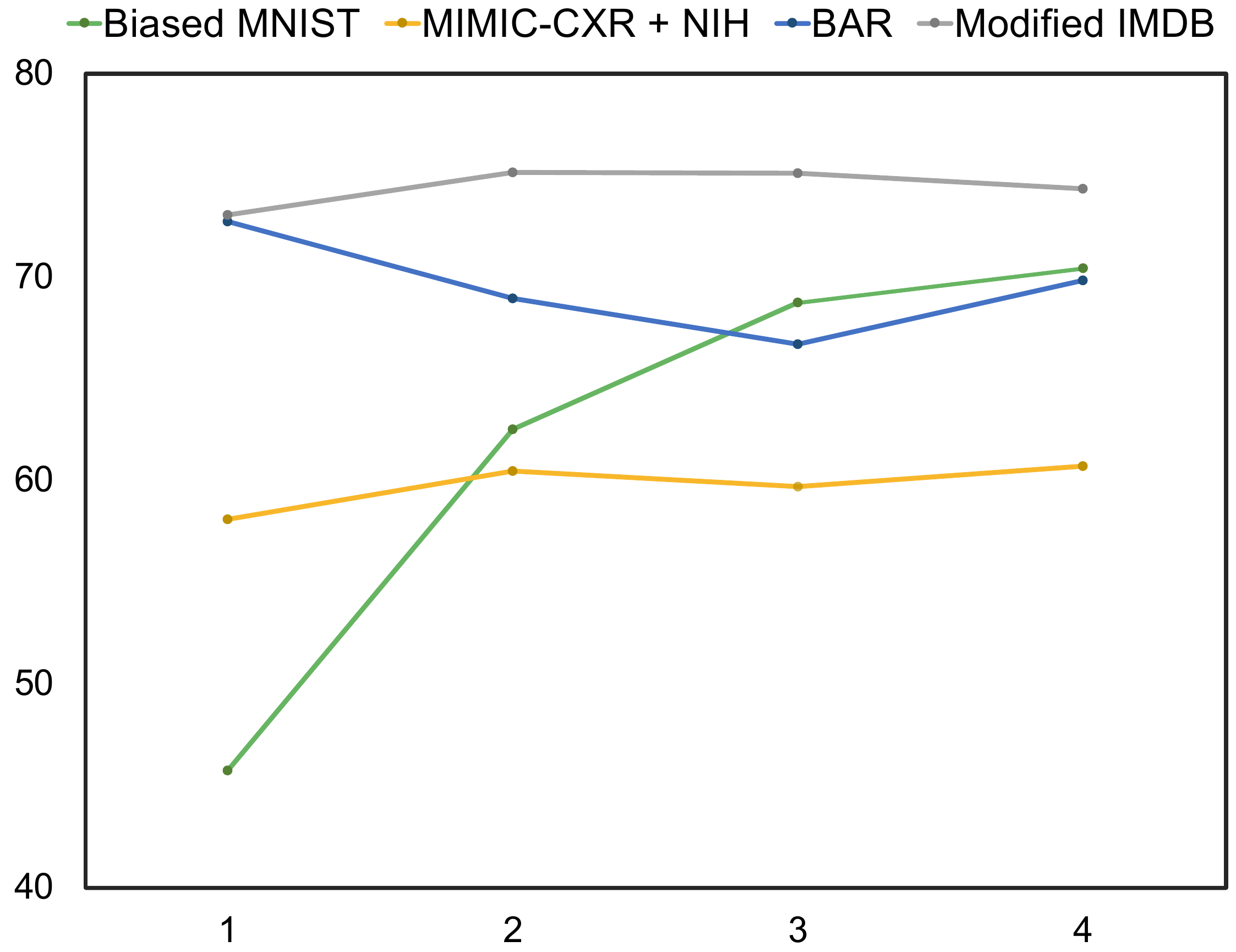}
\end{center}
\caption{The accuracy scores (\%) when changing the number of experts from 1 to 4 
on Biased MNIST, Modified IMDB, MIMIC-CXR + NIH, and BAR datasets.
In the case of multiple biases (Biased
MNIST), the debiased classification accuracy is raised as the number
of experts increases. But for two biases (Modified IMDB) or
single bias (BAR and MIMIC-CXR + NIH), the number of experts
does not affect the performance too much.}
\label{fig_num_blocks}
\end{figure}

\noindent
\textbf{Ablation study for multiple biases-specific experts.}
We individually removed the expert modules inserted into the shallowest block of ResNet-18, to obtain the classification results when the number of experts ranged from 1 to 4 (Fig.~\ref{fig_num_blocks}). 
When multiple biases exist, we can see that the greater the number of blocks covered by the expert, the better the debiased classification effect. 
When the number of biases is two or one, the performance remains almost stable. 
This also confirmed the conclusion from the exploratory experiment and the sufficiency of our strategy.

\begin{table}[tb]
{\centering
\caption{Further ablation study for multiple biases-specific experts on Biased MNIST and MIMIC-CXR + NIH. The accuracy (\%) results reveal that inserting multiple biases-specific experts at different depths of PnD is effective.} 
\label{ablation_ensemble}
\resizebox{\linewidth}{!}{
\begin{tabular}{cccccc}
\toprule

\multicolumn{4}{c}{Method} & \multicolumn{1}{c}{Biased MNIST} & \multicolumn{1}{c}{MIMIC-CXR + NIH} \\ 
\midrule
\multicolumn{4}{l}{MoE on the last block}    &52.06 $\pm$1.10  &58.83 $\pm$ 0.42  \\
\multicolumn{4}{l}{MoE on all blocks w/o other debiasing strategies} &67.49 $\pm$1.41     & 58.47  $\pm$ 0.45  \\
\midrule

\multicolumn{4}{l}{PnD}      & 70.43 $\pm$ 0.74     & 60.73  $\pm$ 0.87  \\ 

\bottomrule
\end{tabular}
}
}
\end{table}

We further evaluate the performance of multiple experts ensemble in Tab.~\ref{ablation_ensemble}.  
We give the results of ensemble multiple biases-specific experts only in the last block (2nd row). When a single bias (MIMIC-CXR
+ NIH) exists, its performance drops by 1.9\% compared to the PnD. However, when multiple biases (Biased MNIST), the performance decreases significantly by 18.37\%. 
It fully illustrates the effectiveness of our idea that we should remove multiple biases from different depths in the network. 
In order to distinguish the effect of MoE and other debiasing strategies, we added an additional set of experiments, where we only keep the classification loss without weight for samples and the gating loss. The results show that the MoE strategy can also achieve relatively great results (3rd row). 
This is because we consider the features from the network in a depth-by-depth manner. The network focuses on more diverse regions, thus outputting a prediction result that is not limited to a single feature, which may be bias.

\subsection{Detailed analysis}

\begin{table*}[]
\centering
\caption{Accuracy scores on worst groups and all groups for wearing lipstick or not classification on CelebA, where bias attributes are attractive or not, heavy makeup or not, high cheekbone or not, and gender. In the last column, we average the results in four bias attributes. The best results are highlighted in \textcolor{blue}{\textbf{blue}}, and the second best results are in \textcolor{red}{\textbf{red}}.}
\label{tab_celeba_results}
\resizebox{1\linewidth}{!}{
\begin{tabular}{l|ll|ll|ll|ll|ll}
\toprule
\multicolumn{1}{c|}{\multirow{2}{*}{Method}} & \multicolumn{2}{c|}{Attractive or not} & \multicolumn{2}{c|}{Heavy makeup or not} & \multicolumn{2}{c|}{High cheekbones or not} & \multicolumn{2}{c|}{Gender} &\multicolumn{2}{c}{Average}  \\
\multicolumn{1}{c|}{} & \multicolumn{1}{c}{Worst group} & \multicolumn{1}{c|}{All groups} & 
\multicolumn{1}{c}{Worst group} & \multicolumn{1}{c|}{All groups} &
\multicolumn{1}{c}{Worst group} & \multicolumn{1}{c|}{All groups} &
\multicolumn{1}{c}{Worst group} & \multicolumn{1}{c|}{All groups} &
\multicolumn{1}{c}{Worst group} & \multicolumn{1}{c}{All groups} \\ 
\midrule
ResNet-18~\cite{he2016deep}&\textcolor{red}{\textbf{85.92}} $\pm$ 0.19 &\textcolor{red}{\textbf{91.53}} $\pm$ 0.29 & \textcolor{red}{\textbf{26.23}} $\pm$ 1.73 &\textcolor{red}{\textbf{74.77}} $\pm$ 0.07  &\textcolor{red}{\textbf{91.54}} $\pm$ 0.48 &\textcolor{red}{\textbf{93.32}} $\pm$ 0.05 &25.96 $\pm$ 4.08 &71.54 $\pm$ 0.99   &57.41 $\pm$ 1.62 &82.79 $\pm$ 0.35  \\
DebiAN~\cite{li2022discover} &85.82 $\pm$ 1.56 &91.49 $\pm$ 0.14 &24.51 $\pm$ 6.24 &74.37 $\pm$ 0.55  &90.27 $\pm$ 2.14 &93.03 $\pm$ 0.26 &\textcolor{red}{\textbf{32.05}} $\pm$ 1.81  &\textcolor{blue}{\textbf{73.41}} $\pm$ 3.33    &\textcolor{red}{\textbf{58.16}} $\pm$ 2.94 &\textcolor{red}{\textbf{83.07}} $\pm$ 1.07   \\
\midrule
PAD  &\textcolor{blue}{\textbf{87.33}} $\pm$ 1.94 &\textcolor{blue}{\textbf{91.77}} $\pm$ 0.02  &\textcolor{blue}{\textbf{27.94}} $\pm$ 6.32  &\textcolor{blue}{\textbf{75.08}} $\pm$ 1.17   &\textcolor{blue}{\textbf{92.00}} $\pm$ 1.29 &\textcolor{blue}{\textbf{93.37}} $\pm$ 0.08 &\textcolor{blue}{\textbf{32.69}} $\pm$ 5.44 &\textcolor{red}{\textbf{73.39}} $\pm$ 0.51  &\textcolor{blue}{\textbf{60.00}} $\pm$ 1.53 &\textcolor{blue}{\textbf{83.40}} $\pm$ 0.44  \\
\bottomrule
\end{tabular}
}
\end{table*}

\noindent
\textbf{More results on real-world dataset.}
We additionally evaluate the performance on CelebA~\cite{liu2015faceattributes}, a real-world dataset, in Tab.~\ref{tab_celeba_results}. For wearing lipstick or not classification, we show the accuracy scores of worst group and all groups in four bias attributes (2nd -- 5th cols). 
From this table, we can see that the performance of PnD outperforms other methods in worst groups and all groups of almost all bias attributes. It demonstrates the advantage of our method in removing agnostic biases for real-world dataset.

\begin{figure}[tb]
\begin{center}
\centering\includegraphics[width=0.47\textwidth]{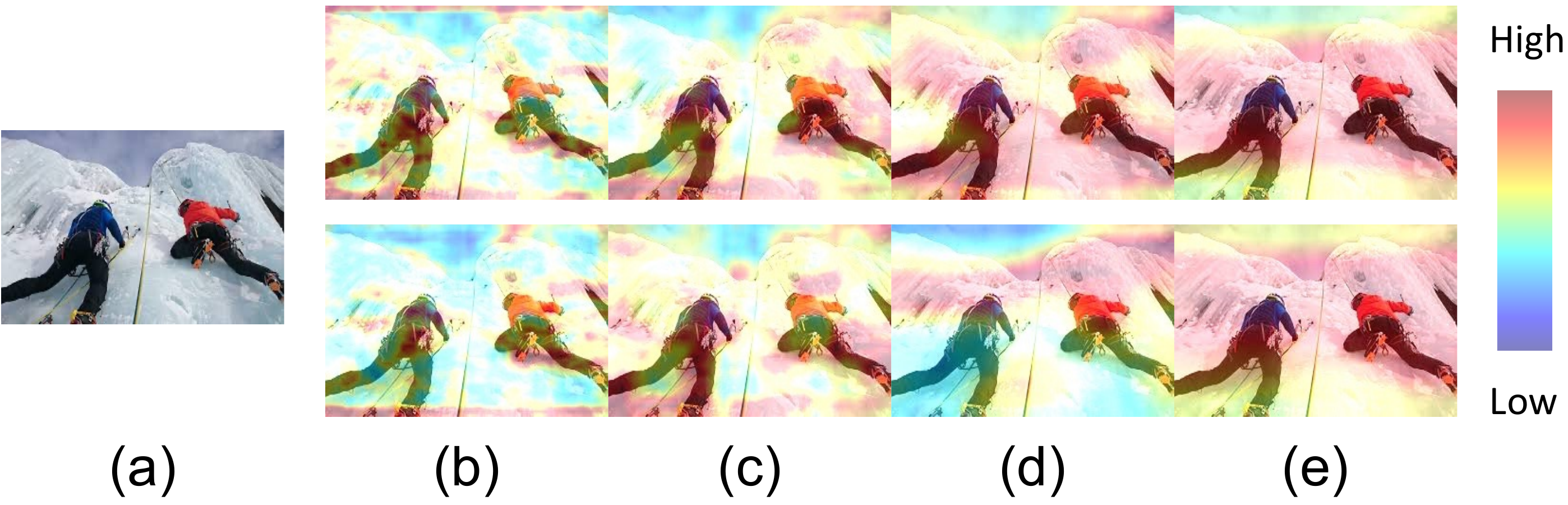}
\end{center}
\caption{Regions of interest (ROIs) for biases-specific experts of our PnD in the debiased (upper) and bias (lower) encoders, when conducting action classification in the test set of BAR. (a) are original images, (b)-(e) are their saliency maps generated using Grad-CAM, from the first expert to the fourth expert. The ROIs for debiased classification and bias detection are changing as the network gets deeper, and there are also significant differences between the two tasks.}
\label{fig_bias_cam}
\end{figure}

\noindent
\textbf{Visualization of learned target and bias features.}
We visualized the region of interest of each expert using Grad-CAM~\cite{selvaraju2017grad} (Fig.~\ref{fig_bias_cam}) to qualitatively verify the debiased classification (upper) and bias detection (lower) performances of each block in PnD. 
In the debiased classification, all experts were able to focus on the target region (climbing). 
Meanwhile, in bias detection, each expert could handle bias features differently; for instance, the first expert focuses on background texture, and the third expert concentrates on snowy slopes.
We conclude that the two encoders can capture target and bias features properly and independently.

\begin{table}[tb]
{\centering
\caption{Accuracy scores (\%) for debiased classification results $\mathbf{\hat{y}}_{\rm d}^{(i)}$ of each expert (from $E^{(1)}$ to $E^{(4)}$), and the probability ${\rm p}^{(i)}$ (in parentheses) assigned by the gating module on Biased MNIST, Modified IMDB, MIMIC-CXR + NIH, and BAR datasets. We can see that each dataset has different trends in classification accuracy across experts.} 
\label{tab_experts_out}
\resizebox{\linewidth}{!}{
\begin{tabular}{lccccc}
\toprule
& \multicolumn{1}{c}{$E^{(1)}$} &
\multicolumn{1}{c}{$E^{(2)}$} & 
\multicolumn{1}{c}{$E^{(3)}$} & 
\multicolumn{1}{c}{$E^{(4)}$} & 
\multicolumn{1}{c}{Final}\\ \midrule
Biased MNIST &48.47 (0.07) &68.75 (0.15) &70.66 (0.46) &68.19 (0.32) &70.43 \\
BAR   &33.44 (0.10)  &49.54 (0.19)  &61.57 (0.18)  &69.83 (0.53) &69.83 \\
Modified IMDB &68.13 (0.04) &72.89 (0.35) &74.71  (0.38) &74.25 (0.22) & 74.34\\
MIMIC-CXR+NIH  &51.77 (0.11) &57.90 (0.30) &60.70 (0.23) &61.00 (0.36) &60.73 \\
\bottomrule
\end{tabular}
}
}
\end{table}

\noindent
\textbf{Analysis on the mixed debiased classification of experts.}
To analyze the effectiveness of mixing different expert results using adaptively gating, we checked the output from each expert and the probability assigned by the gating module. 
Tab.~\ref{tab_experts_out} shows that, across these four datasets, the highest debiased classification accuracy scores were located at 3rd or 4th expert due to the different complexities of biases.
Additionally, the results of the expert modules slightly exceeded the final results coordinated by multiple experts. 
This is reasonable because shallow blocks may contain few target features, resulting in poor performance of the corresponding experts, and thus degrading the final results. This occurs particularly when the probability ${\rm p}^{(i)}$ assigned to the best expert is relatively low. However, we cannot remove shallow experts directly (see the results of reducing the number of experts in Fig.~\ref{fig_num_blocks}). We may be able to enhance the performance by selecting the output of the expert that has the highest ${\rm p}^{(i)}$ during testing.

\noindent 
\textbf{Limitations.}
We employed multiple expert modules, which inevitably increased the number of network parameters.
One potential solution is to increase the sparsity of expert networks ~\cite{fedus2022switch}. Furthermore, as discussed in~\cite{li2022WhacAMole}, removing multiple unknown biases without an inductive bias is difficult. Due to the requirement of unbiased training data, PnD does not perform as well on fully biased datasets such as BAR. We will examine these issues further in the future.

\noindent 
\textbf{Societal impacts.}
This study introduces a more realistic bias scenario and provides a simple yet effective approach to ensure that deep learning-based decision processes are not biased toward agnostic attributes in the data.
We believe that the proposed approach will encourage the development of more trustworthy AI applications. 
For example, it can increase racial and gender equity in face recognition systems by protecting minority populations from systemic biases.

\section{Conclusion}

Existing bias mitigation methods struggle to deal with multiple unknown biases in real-world scenarios.
To address these limitations, we presented a novel bias scenario, namely, agnostic biases mitigation. 
First, we investigated our hypothesis that different bias features would cluster at different depths in a network. 
We then proposed a PnD method to address the new scenario by dividing the bias space into multiple subspaces across network depths and removing them using a mixture of biases-specific experts. Extensive experiments on both public and our constructed datasets demonstrated PnD's excellent performance.

\noindent
\textbf{Acknowledgement.}
This work was supported by JST SPRING Grant Number JPMJSP2108, Institute for AI and Beyond of the University of Tokyo, JSPS KAKENHI Grant Numbers JP23H03449, JP23KJ0404, and JP22K17947.

{\small
\bibliographystyle{ieee_fullname}
\bibliography{egbib}
}

\newpage
\setcounter{section}{0}
\setcounter{figure}{0}
\setcounter{table}{0}

\input{supp.tex}

\end{document}

%% file: supp.tex
\renewcommand\thesection{\Alph{section}}
\renewcommand\thefigure{\Alph{figure}} 
\renewcommand\thetable{\Alph{table}}

\newcommand{\vmduc}[1]{\textcolor{red}{#1}}
\newcommand{\li}[1]{\textcolor{blue}{#1}}

\iccvfinalcopy 

\def\iccvPaperID{9141} 
\def\httilde{\mbox{\tt\raisebox{-.5ex}{\symbol{126}}}}





\twocolumn[
\centering
\section*{
\Large
Supplementary Material for \\Partition-and-Debias: Agnostic Biases Mitigation via A Mixture of Biases-Specific Experts}
\vspace{0.5cm}
]


This supplementary material complements our paper with the following parts: First, we present details in implementing our PnD (Sec.~\ref{sec_imple}), which are not included in the main paper. Second, we provide more details on the dataset used in the main paper and some concepts related to biases (Sec.~\ref{sec_data}). Finally, we add more analysis to assess PnD (Sec.~\ref{sec_add}).


\begin{table}[tb]
\centering
\caption{The main network structure of PnD. We omit BatchNorm and ReLU operations in this table. Although PnD has four debiased blocks, four bias blocks, and four biases-specific experts with a debiased classifier and a bias classifier. We only show the debiased/bias parts here since the two parts have the same structure.}
\label{network}
\resizebox{1.0\linewidth}{!}{
\begin{tabular}{ccc}
\toprule
Layer name &Intput size &Operation \\ 
\midrule
Initial layer &$160\times 160\times 3$ &$[7\times 7, 64]\times 1$ conv\\ 
\midrule
Initial pooling &$80\times 80\times 64$  & $[3\times 3]$ maxpool \\ \midrule
\multirow{2}{*}{Block 1}  & \multirow{2}{*}{$40\times 40\times 64$} & \multirow{2}{*}{$\begin{bmatrix} 3\times 3, 64 \\ 3\times 3, 64 \end{bmatrix}\times 2$ conv} \\
&  &    \\
\midrule
\multirow{5}{*}{Expert 1} & \multirow{5}{*}{$40\times 40\times 64$}   & \multirow{2}{*}{$\begin{bmatrix} 3\times 3, 128 \\ 3\times 3, 512 \end{bmatrix}\times 1$ conv}\\
&  &    \\\cmidrule{3-3}
& & \multirow{1}{*}{adaptive avgpool} \\\cmidrule{3-3}
& & \multirow{2}{*}{$\begin{bmatrix} 1024, 16\\ 16, 10\end{bmatrix}\times 1$ linear} \\
&  &   \\
\midrule
\multirow{2}{*}{Block 2}  & \multirow{2}{*}{$40\times 40\times 64$}  & \multirow{2}{*}{$\begin{bmatrix} 3\times 3, 128 \\ 3\times 3, 128 \end{bmatrix}\times 2$ conv} \\
&  & \\
\midrule
\multirow{5}{*}{Expert 2} & \multirow{5}{*}{$20\times 20\times 128$}   & \multirow{2}{*}{$\begin{bmatrix} 3\times 3, 128 \\ 3\times 3, 512 \end{bmatrix}\times 1$ conv}\\
&  &    \\\cmidrule{3-3}
& & \multirow{1}{*}{adaptive avgpool} \\\cmidrule{3-3}
& & \multirow{2}{*}{$\begin{bmatrix} 1024, 16\\ 16, 10\end{bmatrix}\times 1$ linear} \\
&  &   \\
\midrule
\multirow{2}{*}{Block 3}  & \multirow{2}{*}{$20\times 20\times 128$}  & \multirow{2}{*}{$\begin{bmatrix} 3\times 3, 256 \\ 3\times 3, 256 \end{bmatrix}\times 2$ conv} \\
&  & \\
\midrule
\multirow{5}{*}{Expert 3} & \multirow{5}{*}{$10\times 10\times 256$}   & \multirow{2}{*}{$\begin{bmatrix} 3\times 3, 128 \\ 3\times 3, 512 \end{bmatrix}\times 1$ conv}\\
&  &    \\\cmidrule{3-3}
& & \multirow{1}{*}{adaptive avgpool} \\\cmidrule{3-3}
& & \multirow{2}{*}{$\begin{bmatrix} 1024, 16\\ 16, 10\end{bmatrix}\times 1$ linear} \\
&  &   \\
\midrule
\multirow{2}{*}{Block 4}  & \multirow{2}{*}{$10\times 10\times 256$}   & \multirow{2}{*}{$\begin{bmatrix} 3\times 3, 512 \\ 3\times 3, 512 \end{bmatrix}\times 2$ conv} \\
&  & \\
\midrule
\multirow{5}{*}{Expert 4} & \multirow{5}{*}{$5\times 5\times 512$}   & \multirow{2}{*}{$\begin{bmatrix} 3\times 3, 128 \\ 3\times 3, 512 \end{bmatrix}\times 1$ conv}\\
&  &    \\\cmidrule{3-3}
& & \multirow{1}{*}{adaptive avgpool} \\\cmidrule{3-3}
& & \multirow{2}{*}{$\begin{bmatrix} 1024, 16\\ 16, 10\end{bmatrix}\times 1$ linear} \\
&  &   \\
\bottomrule
\end{tabular}
 }
\end{table}

\section{Implementation Details}
\label{sec_imple}
\noindent
\textbf{Network structure.} The main network structure is shown in Tab.~\ref{network}, including details in layer names, input sizes and operations. Experts 1 -- 4 are the inserted biases-specific experts in PnD, the remainings denote the layers and blocks of the debiased/bias encoder in it, which is the same with ResNet-18~\cite{he2016deep}. In the third operation of each expert, the linear layer input dimension is the doubled dimension from adaptive avgpool, since it will take the concatenation of debiased features and bias features as the input. The BatchNorm and ReLU operations are not shown in this table for simplicity. Only one kind of  debiased/bias encoder is given in the table because their structures are the same.

\noindent
\textbf{Diversity loss.}
The $\textsc{KL}\left(\mathbf{\hat{y}}_{\rm b}^{(i)}, \mathbf{\hat{y}}_{\rm b}^{(i-1)}\right)$ in diversity loss $\mathcal{L}_{\textup{div}}$ is calculated as:

\begin{equation}
\textsc{KL}\left(\mathbf{\hat{y}}_{\rm b}^{(i)}, \mathbf{\hat{y}}_{\rm b}^{(i-1)}\right)= \mathbf{\hat{y}}_{\rm b}^{(i)} \log \left(\frac{\mathbf{\hat{y}}_{\rm b}^{(i)}}{\mathbf{\hat{y}}_{\rm b}^{(i-1)}}\right), \nonumber
\end{equation}

\noindent
where $\mathbf{\hat{y}}_{\rm b}^{(i)}$ denotes the bias prediction distribution from $i^{th}$ expert, $\mathbf{\hat{y}}_{\rm b}^{(i-1)}$ denotes that from the previous one.

\noindent
\textbf{Training details.} In all experiments, we train our framework with two-phase optimization using Adam with a batch size of 128. Following settings in \cite{nam2020learning}, the hyperparameter $q$ in $\textsc{GCE}$ loss is set to 0.7. 

For \{Biased MNIST, BAR, Modified IMDB, MIMIC-CXR + NIH\}, during the initial and counterfactual training stages, we set the epoch number as \{70, 70, 20, 10\} and \{100, 70, 30, 50\}, $\alpha$ as \{0.2, 0.6, 0.2, 0.2\} and \{2.0, 1.0, 2.0, 2.0\}, learning rate (LR) as \{1e-3, 1e-4, 5e-4, 5e-4\} and \{5e-4, 5e-5, 5e-4, 1e-4\}, LR is decent every \{20, -, -, -\}, and \{20, 10, -, 30\} epochs with an LR Decay Gamma of 0.5, respectively. In the second stage, $\beta$ is set as \{4.0, 0.3, 0.3, 0.3\}, $K$ is 16, $P$ is 8,  and the temperature hyperparameter in contrastive loss is \{0.1, 0.07, 0.1, 0.1\}. In both two stages, weight decay is \{1e-5, 5e-6, 1e-6, 1e-6\}. We do not change any hyperparameters when the bias ratio is different for a certain dataset.
``-" means no specific value.

For additional experiments on CelebA, during the initial and counterfactual training stages of PnD, we set the epoch number as 10 and 20, $\alpha$ as 0.2 and 2.0, respectively. In both two stages, learning rate (LR) is 5e-4, weight decay is 1e-6. In the second stage, $\beta$ is set as 0.3, $K$ is 16, $P$ is 8, LR is decent every 10 epochs with an LR Decay Gamma of 0.5.

\section{Datasets}
\label{sec_data}
\subsection{Public datasets}

\begin{figure}[t]
\begin{center}
\centering\includegraphics[width=0.45\textwidth]{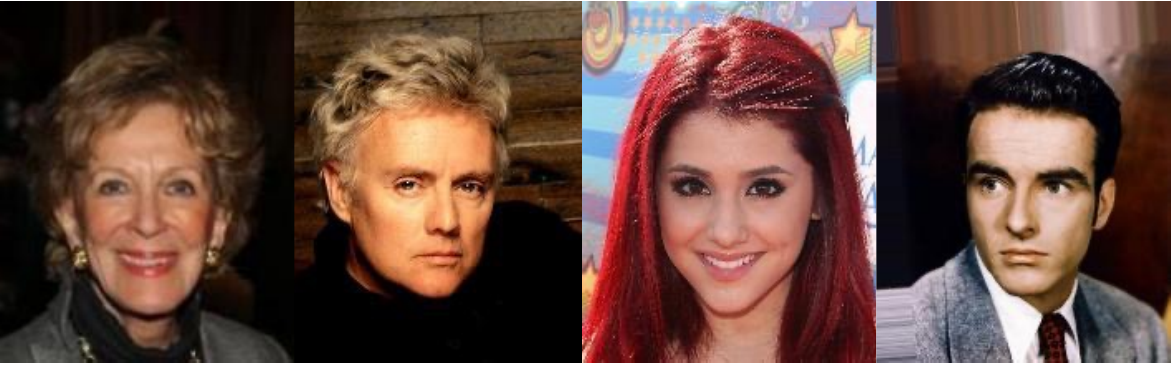}
\end{center}
\caption{Examples from CelebA dataset.}
\label{fig_celeba}
\end{figure}

\begin{figure}[t]
\begin{center}
\centering\includegraphics[width=0.46\textwidth]{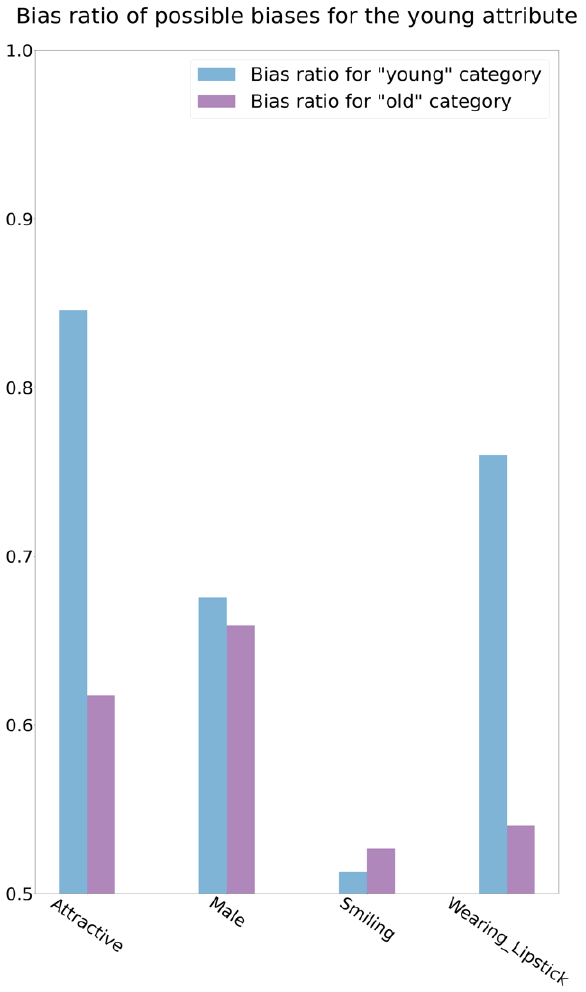}
\end{center}
\caption{Bias ratio of all possible biases for the age attribute in CelebA.}
\label{fig_young}
\end{figure}

\begin{figure*}[t]
\begin{center}
\centering\includegraphics[width=\textwidth]{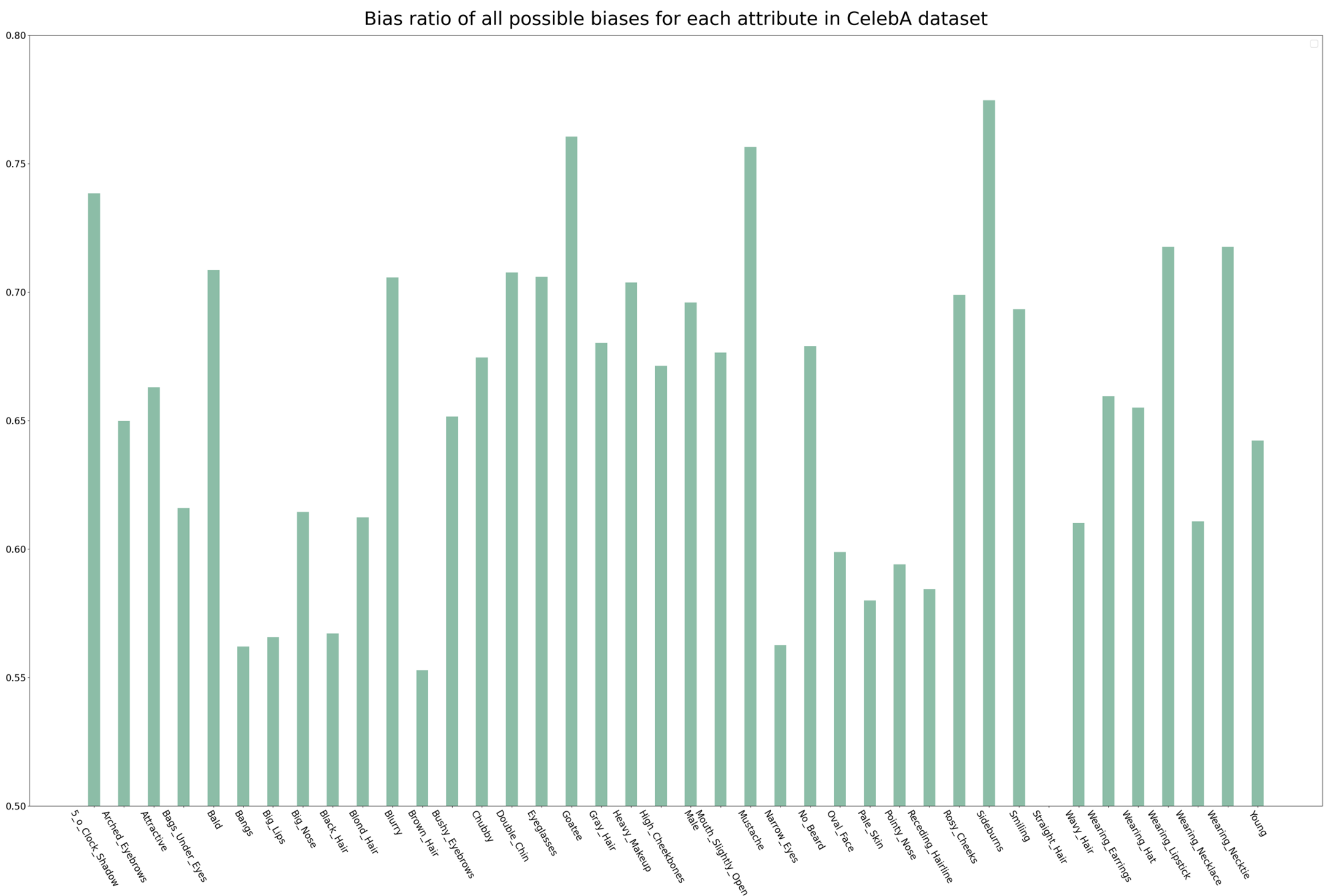}
\end{center}
\caption{Bias ratio of all possible biases for each attribute in CelebA dataset. We can find that most attributes are biased. Note that in this figure (also in other figures), we use the original annotation from CeleA as the name of attribute.} 
\label{fig_celeba_bias_dis}
\end{figure*}

\noindent
\textbf{CelebA} ~\cite{liu2015faceattributes}. It is a publicly available face attribute dataset that contains 202599 face images of 10177 celebrity identities, each with 40 attribute annotations (Fig.~\ref{fig_celeba}). 

Take the age attribute for example, if most images with \textit{young} category are annotated with \textit{female}, while most images with \textit{old} category are annotated with \textit{male}. We can consider the gender attribute as a bias for age attribute. In this bias scenario, we have two concepts relevant to the bias problem:

\noindent
(i) Bias ratio. It denotes the probability of co-occurrence of the bias category and the target category. For \textit{young} in age attribute, the bias ratio of \textit{female} in gender bias attribute refers to the proportion of individuals with both \textit{female} and \textit{young} in the total number of individuals with \textit{young}.

\noindent
(ii) Worst group. Following the definition in \cite{li2022discover}, the worst group in this paper denotes the group that gets the lowest accuracy score among all 4 combinations of the target categories and the bias categories, such as (\textit{young}, \textit{old}) $\times$ (\textit{male}, \textit {female}).

As shown in Fig.~\ref{fig_celeba_bias_dis}, we analyze this dataset by calculating the bias ratio of possible biases for each attribute in this dataset. That is, for each target attribute, we analyze the percentage of other attributes within each of its categories (here, two categories), and if certain other attribute would be a bias as described above, we calculate the bias ratio for each category of the target attribute as in Fig.~\ref{fig_young}. For each target attribute, all bias ratios of possible biases are averaged. From Fig.~\ref{fig_celeba_bias_dis}, we can see most attributes are biased in CelebA.

\begin{figure}[t]
\begin{center}
\centering\includegraphics[width=0.47\textwidth]{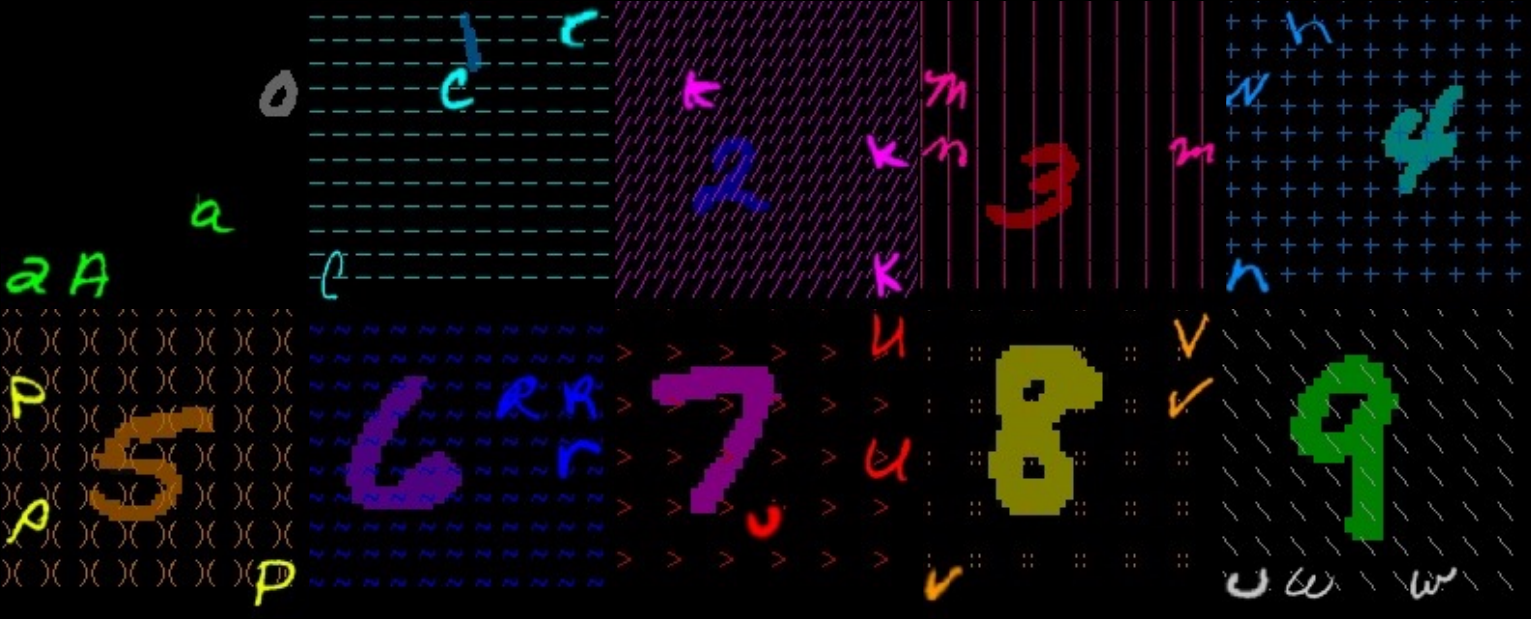}
\end{center}
\caption{Examples for digit \textit{0} -- \textit{9} from Biased MNIST training data, there are 7 biases in this dataset, including digit color, digit scale, digit position,
type of background texture, background texture color, co-occurring letter, and letter color.}
\label{fig_bmnist}
\end{figure}

\noindent
\textbf{Biased MNIST}~\cite{shrestha2022occamnets}. It contains 10 digits (\textit{0} -- \textit{9}) as its target categories and 7 biases: digit color, digit scale, digit position, type of background texture, background texture color, co-occurring letter, and letter color (Fig.~\ref{fig_bmnist}).
There are 50000, 10000, and 10000 images for training, validation, and test.

\begin{figure}[t]
\begin{center}
\centering\includegraphics[width=0.46\textwidth]{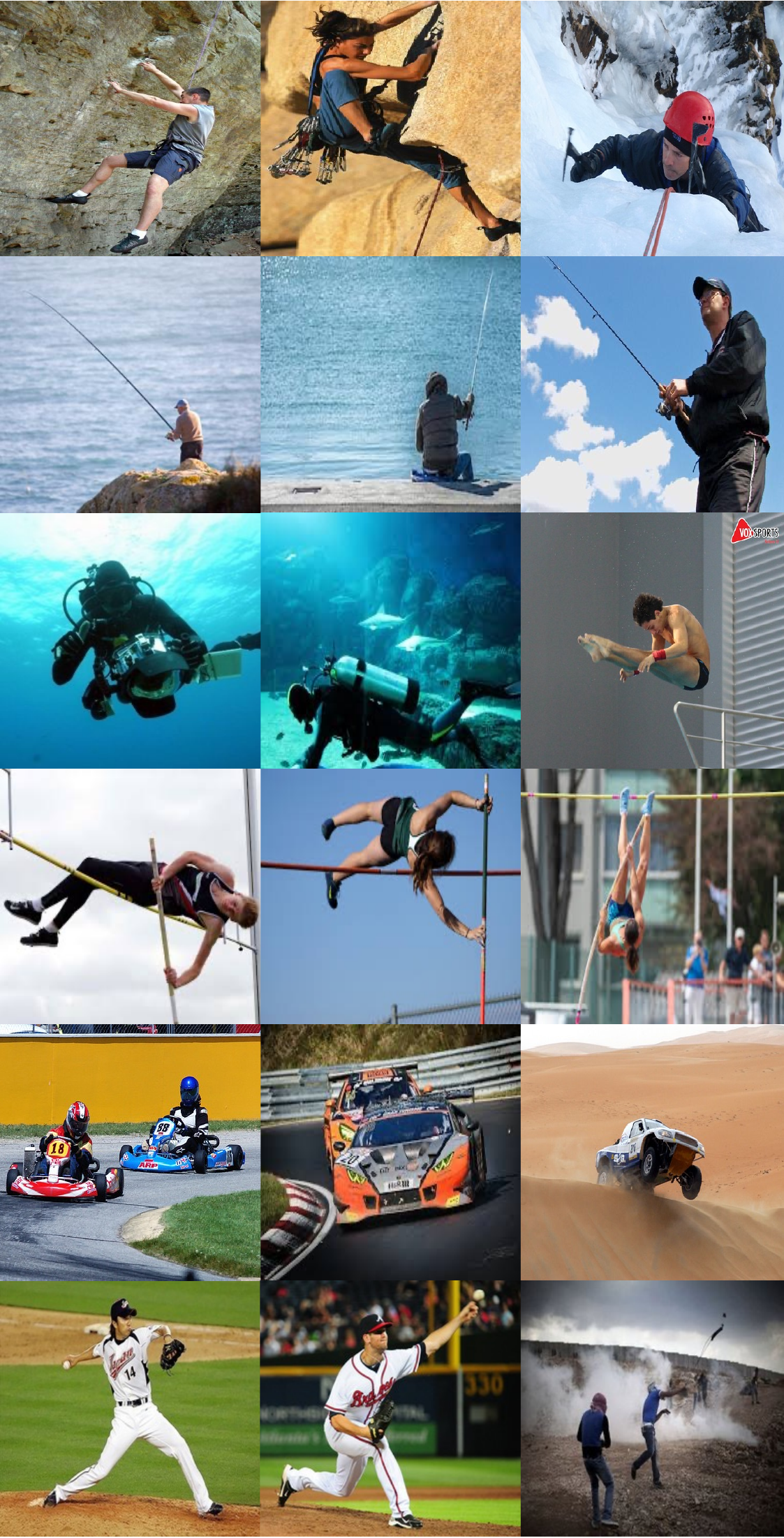}
\end{center}
\caption{Examples from BAR dataset. The images in 1st and 2nd cols are training data with action and corresponding co-occurring background pairs, including climbing and rockwall, fishing and water surface, diving and underwater, vaulting and sky, racing and a paved track, throwing and playing field (from 1st -- 6th rows). The images in 3rd col are test data, all of which are scenarios not seen in the training set. }
\label{fig_bar}
\end{figure}

\noindent
\textbf{BAR}~\cite{nam2020learning}. There are typical action-place pairs, including \textit{climbing} and \textit{rockwall},  \textit{fishing} and \textit{water surface},  \textit{diving} and \textit{underwater},  \textit{vaulting} and \textit{sky}, \textit{racing} and \textit{a paved track}, \textit{throwing} and \textit{playing field} in the 1941 training images (Fig.~\ref{fig_bar}, 1st and 2nd cols); and unseen samples beyond the settled pairs in the 654 test images (Fig.~\ref{fig_bar}, 3rd col). BAR can be seen as a dataset with a single bias, where the action is spuriously correlated with the background.

\subsection{Our contructed datasets}

\begin{figure}[t]
\begin{center}
\centering\includegraphics[width=0.47\textwidth]{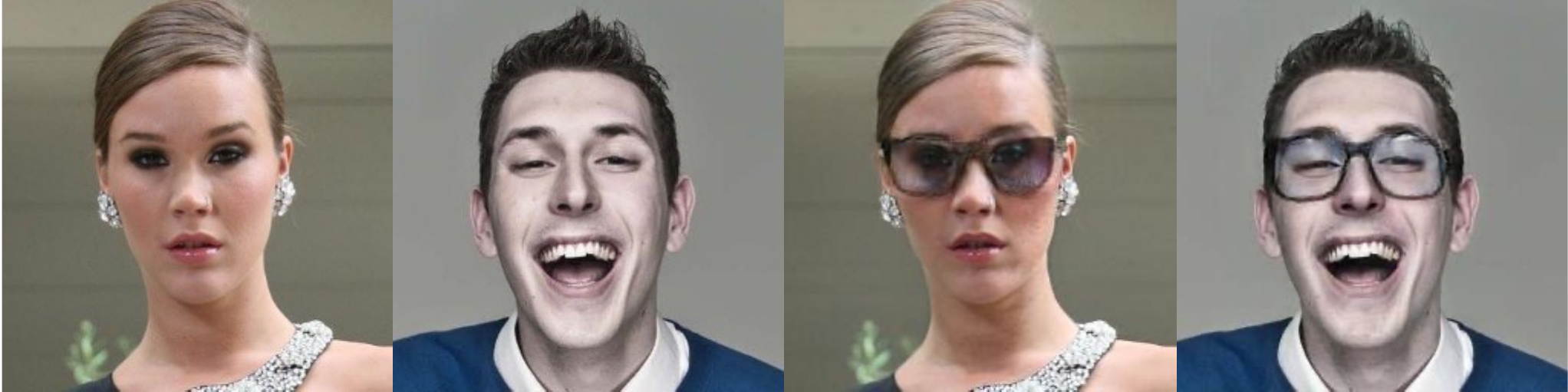}
\end{center}
\caption{The orginal images (left two) from IMDB and the images with glasses (right two) generated by a face attribute transfer model.}
\label{fig_wear_glass}
\end{figure}

\noindent
\textbf{Modified IMDB.} Original IMDB~\cite{Rasmus_2018_Deep} contains 460723 face images from 20284 celebrities. We use the cleaned IMDB dataset with 112340 face images provided by \cite{kim2019learning}, which is filtered by pretrained models designed for age and gender classification. Their cleaned IMDB consists of extreme bias 1 (EB1), extreme bias 2 (EB2), and a test set. EB1 contains 36004 face images, which are old people (aged 40+) and men, or are young people (aged 0 -- 29) and women. EB2 contains 16800 face images, which are opposite to EB1. The test set contains 13129 images, all are aged 0 -- 29 or 40+ without other settings. The gender attribute is a bias for the age classification in the cleaned IMDB dataset. In order to add another natural bias to this dataset, so that it has two biases for age classification. As shown in Fig.~\ref{fig_wear_glass}, we use a face attribute transfer model~\cite{guo2021image} to put glasses on the faces. Thus, we can make \textit{wearing glasses} also a bias in this dataset by controlling the ratio of \textit{wearing glasses}. As a result, we create Modified IMDB, where most images \textit{young} labels are \textit{female} and \textit{wearing glasses} and most images with \textit{old} labels are \textit{male} and \textit{not wearing glasses}.
It consists of 20000 training images with a bias ratio of 0.95 and 0.99, 1617 unbiased validation images, and 1617 unbiased test images.

\begin{figure}[t]
\begin{center}
\centering\includegraphics[width=0.47\textwidth]{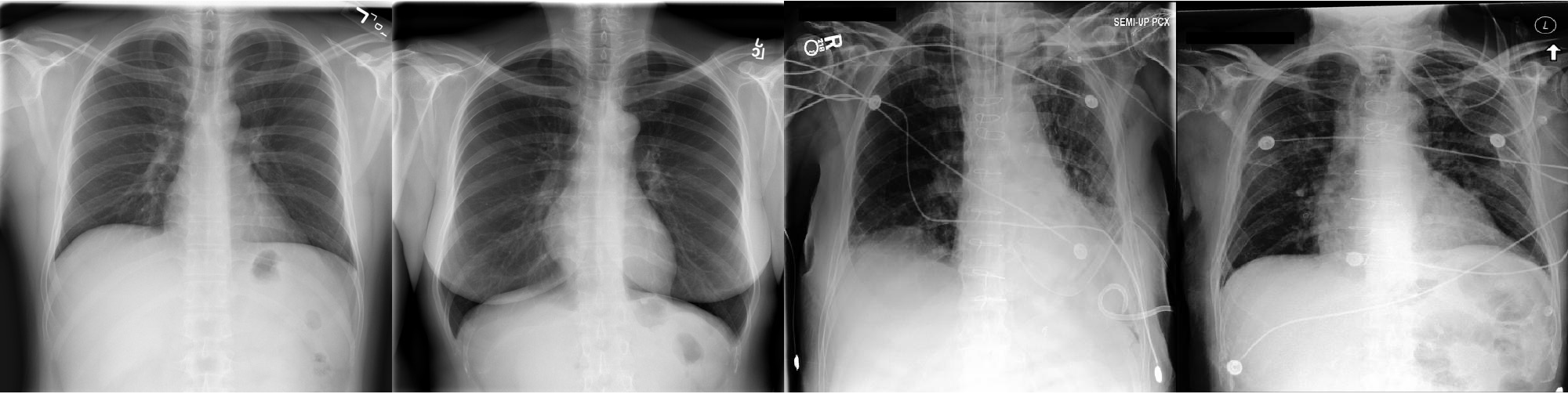}
\end{center}
\caption{Examples from MIMIC-CXR + NIH dataset. They are source-biased, the left two are from NIH with \textit{no finding} labels, and the right two are from MIMIC-CXR with \textit{pneumonia} labels.}
\label{fig_obmed}
\end{figure}

\noindent
\textbf{MIMIC-CXR + NIH.} It is constructed by simulating biases brought by different data sources when collecting datasets. We mix MIMIC-CXR~\cite{Johnson_2019_MIMIC} and NIH~\cite{Wang_2017_Chestx} datasets into a MIMIC-CXR + NIH dataset. The original NIH contains 50500 \textit{no finding} and 876 \textit{pneumonia} training images, 9861 \textit{no finding} and 555 \textit{pneumonia} test images. The original MIMIC-CXR has 10145 \textit{no finding} and 7209 \textit{pneumonia} training images, 122 \textit{no finding} and 140 \textit{pneumonia} test images. Considering \textit{pneumonia} images are very few in NIH, we construct MIMIC-CXR + NIH by collecting most \textit{pneumonia} images from MIMIC-CXR, while most \textit{no finding} images from NIH (Fig~\ref{fig_obmed}). In MIMIC-CXR + NIH, the target categories are \textit{no finding} and \textit{pneumonia}, and the biases come from two data sources. It contains 8500 training images with a bias ratio of 0.80 and 0.95, 500 unbiased validation images, and 500 unbiased test images.

\begin{figure}[t]
\begin{center}
\centering\includegraphics[width=0.47\textwidth]{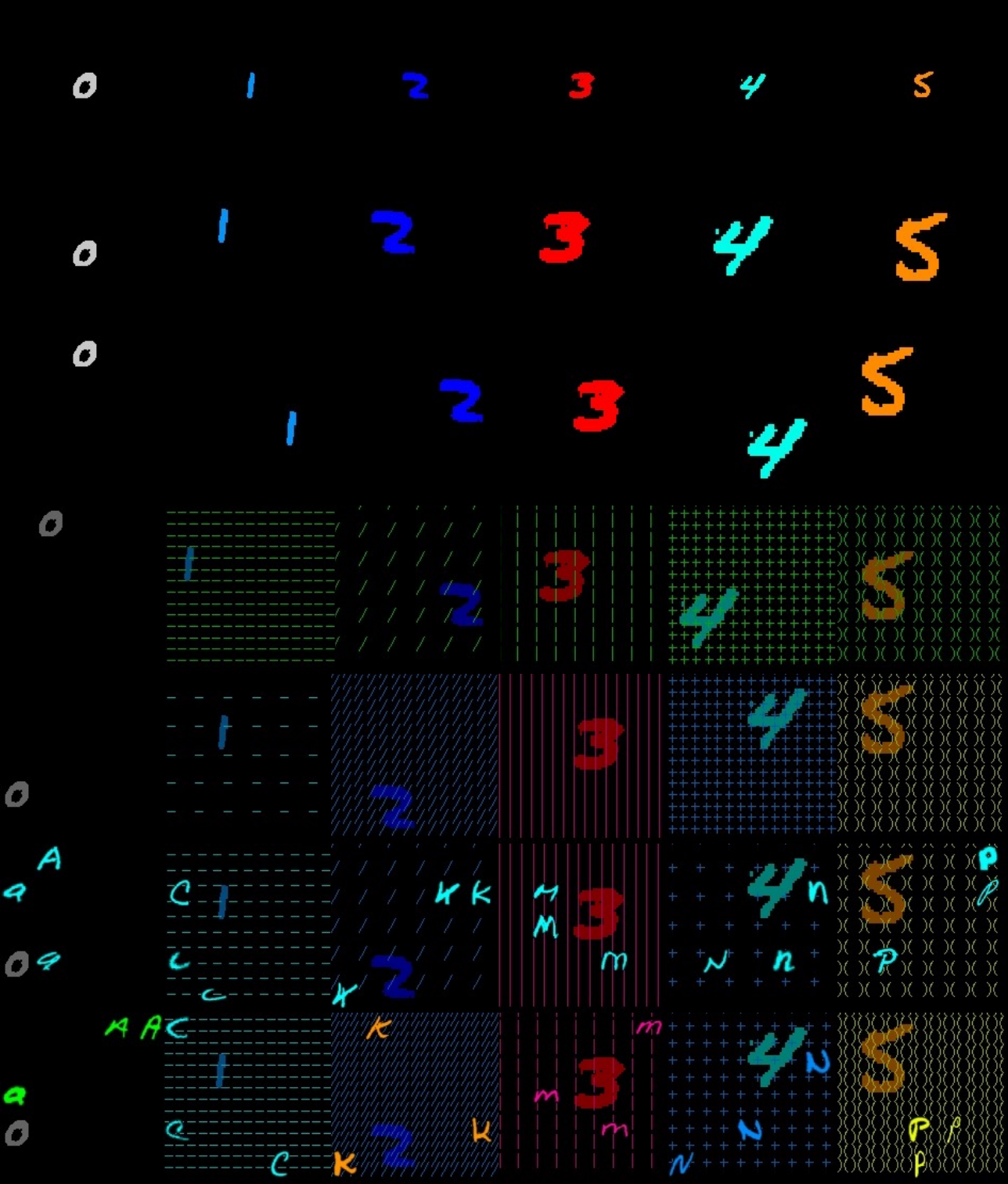}
\end{center}
\caption{Examples from multiple Biased MNISTs with a different number of biases, 1st -- 7th rows indicate the number from 1 -- 7 (gradually adding digit color, digit scale, digit position, texture, texture color, letter, and letter color biases). Note that we only show the same samples with 0 -- 5 digits here in all cases for clarity, we have 10 digits in total.}
\label{fig_mul_bmnist}
\end{figure}

\noindent
\textbf{Multiple Biased MNISTs.} This set of datasets is created according to the method in constructing Biased MNIST~\cite{shrestha2022occamnets}. It consists of 7 Biased MNIST datasets with different numbers (ranging from 1 to 7) of biases. As shown in Fig.~\ref{fig_mul_bmnist}, we construct this set of multiple Biased MNISTs by gradually adding digit color, digit scale, digit position, texture, texture
color, letter, and letter color biases (from 1st -- 7th rows) into MNIST~\cite{lecun2010mnist}. 

\section{Additional Detailed Analysis}
\label{sec_add}

\begin{figure}[t]
\begin{center}
\centering\includegraphics[width=0.46\textwidth]{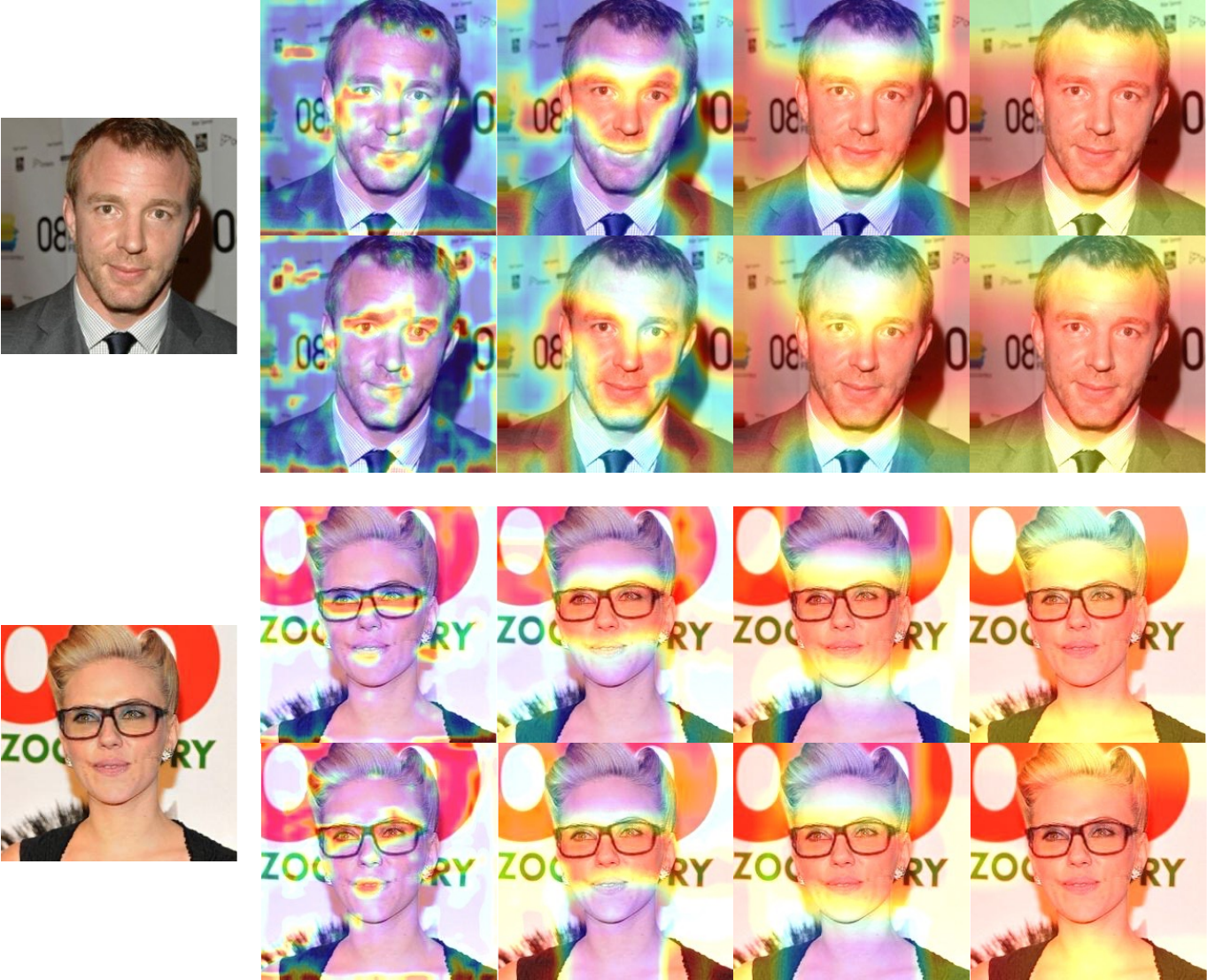}
\end{center}
\caption{Regions of interest for biases-specific experts of our PnD in the debiased (1st and 3rd rows) and bias (2nd and 4th rows) encoders, when conducting action classification in the test set of Modified IMDB. The original images are in the 1st column, and 2nd -- 5th columns are their saliency maps generated using Grad-CAM, from the first expert to the fourth expert. The regions of interest for debiased classification and bias detection are changing as the network gets deeper, and there are also significant differences between the two tasks.}
\label{fig_cam}
\end{figure}

\noindent
\textbf{Visualization for learned target and bias features.} In Fig~\ref{fig_cam}, we visualize the region of interest on more examples from the Modified IMDB dataset. The upper one is an image of a young male without glasses, which conflicts with the bias samples (\textit{young}, \textit{female}, \textit{wearing glasses}) in the training set. The lower one is aligned with bias samples. We can see both debiased age classification and bias detection focus on varying level features in different depths. At the same time, there are differences between these two parts. In the first expert's results, debiased age classification is more related to the position under the eyes, which may be related to age. In contrast, bis detection concentrates on the glasses region.